\documentclass[final]{cvpr}
\usepackage[latin9]{inputenc}
\usepackage{array}
\usepackage{float}
\usepackage{url}
\usepackage{multirow}
\usepackage{amsmath}
\usepackage{amssymb}
\usepackage{graphicx}
\usepackage{tablefootnote}
\usepackage{wasysym}
\usepackage[unicode=true,
 bookmarks=false,
 breaklinks=true,pdfborder={0 0 1},backref=page,colorlinks=false]
 {hyperref}
\hypersetup{
 pagebackref=true,colorlinks}

\makeatletter

\providecommand{\tabularnewline}{\\}
\floatstyle{ruled}
\newfloat{algorithm}{tbp}{loa}
\providecommand{\algorithmname}{Algorithm}
\floatname{algorithm}{\protect\algorithmname}

\usepackage{times}
\usepackage{epsfig}

\usepackage{caption}
\usepackage{algorithm}
\usepackage{algpseudocode}

\def\cvprPaperID{7921} 
\def\confYear{CVPR 2021}

\makeatother

\begin{document}
\title{Joint-DetNAS: Upgrade Your Detector with NAS, Pruning and Dynamic
Distillation}
\author{Lewei Yao\\
 Huawei Noah' Ark Lab}
\author{Lewei Yao$^{1}$\thanks{Equal contribution} \hspace{4mm}Renjie Pi$^{1*}$\hspace{4mm}Hang
Xu$^{2}$\thanks{Corresponding author:\textit{ }xbjxh@live.com}\hspace{4mm}Wei
Zhang$^{2}$ \hspace{4mm}Zhenguo Li$^{2}$\hspace{4mm}Tong Zhang$^{1}$\\
$^{1}$Hong Kong University of Science and Technology\hspace{4mm}$^{2}$Huawei
Noah's Ark Lab}
\maketitle
\begin{abstract}
We propose Joint-DetNAS, a unified NAS framework for object detection,
which integrates 3 key components: Neural Architecture Search, pruning,
and Knowledge Distillation. Instead of naively pipelining these techniques,
our Joint-DetNAS optimizes them jointly. The algorithm consists of
two core processes: \textbf{student morphism }optimizes the student's
architecture and removes the redundant parameters, while\textbf{ dynamic
distillation }aims to find the optimal matching teacher. For \textbf{student
morphism}, weight inheritance strategy is adopted, allowing the student
to flexibly update its architecture while fully utilize the predecessor's
weights, which considerably accelerates the search; To facilitate
\textbf{dynamic distillation}, an elastic teacher pool is trained
via integrated progressive shrinking strategy, from which teacher
detectors can be sampled without additional cost in subsequent searches.
Given a base detector as the input, our algorithm directly outputs
the derived student detector with high performance without additional
training. Experiments demonstrate that our Joint-DetNAS outperforms
the naive pipelining approach by a great margin. Given a classic R101-FPN
as the base detector, Joint-DetNAS is able to boost its mAP from 41.4
to 43.9 on MS COCO and reduce the latency by 47\%, which is on par
with the SOTA EfficientDet while requiring less search cost. We hope
our proposed method can provide the community with a new way of jointly
optimizing NAS, KD and pruning.

\thispagestyle{empty} 
\end{abstract}
\vspace{-4mm}

\section{Introduction}

\vspace{-2mm}

\begin{figure}
\begin{centering}
\vspace{-2mm}
\includegraphics[scale=0.6]{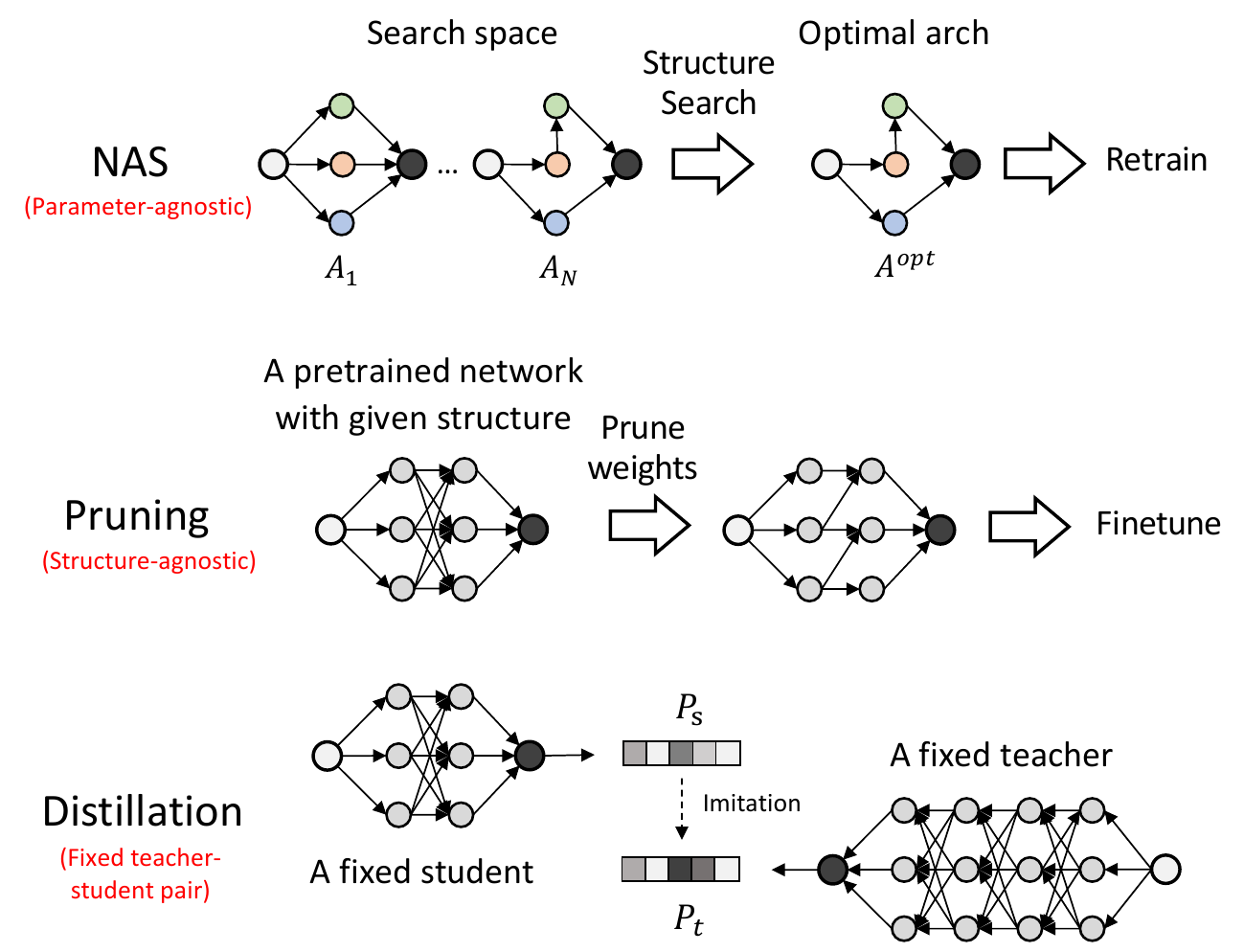}
\par\end{centering}
\vspace{-2mm}

\caption{\label{fig:motivation}Limitation of NAS, pruning and KD. NAS is parameter-agnostic:
The model search and training processes are decoupled, the searched
architecture is retrained from scratch; Pruning is structure-agnostic:
the pre-trained model has a fixed architecture; KD transfers knowledge
between a fixed student-teacher pair while neglecting the structural
dependence between the student and the teacher. Our work aims to jointly
optimize all three methods.}

\vspace{-4mm}
\end{figure}

Finding the optimal tradeoff between model performance and complexity
has always been a core problem for the community. The mainstream approaches
aiming at addressing this issue are: Neural Architecture Search (NAS)
\cite{Ghiasi_2019_CVPR,chen2019detnas,yao2019sm,jiang2020sp} is proposed
to automatically search for promising model architectures; pruning
\cite{li2016pruning,Li2019a,molchanov2016pruning} removes redundant
parameters from a model while maintaining its performance; and Knowledge
Distillation (KD) \cite{hinton2015distilling,chen2017learning,Li_2017_CVPR,wang2019distilling,cho2019efficacy}
aims to transfer the learnt knowledge from a cumbersome teacher model
to a more compact student model. These methods share the same ultimate
goal: boosting the model's performance while making it more compact.
However, jointly optimizing them is a challenging task, especially
for detection, which is much more complex than classification. In
this paper, we propose Joint-DetNAS, a unified framework for detection
which jointly optimizes NAS, pruning and KD.

The aforementioned methods each have some limitations, as illustrated
in Figure \ref{fig:motivation}. NAS and pruning only focus on one
aspect while neglecting the other: The current de facto paradigm of
NAS considers the architecture to be the sole factor that impacts
the model's performance, while pruning only takes parameters into
account and is structure-agnostic. A recent work \cite{frankle2018lottery}
has observed an interesting phenomenon: the pruned model's final performance
highly depends on its retraining initialization. This observation
indicates that the architecture and parameters are closely coupled
with each other, both of them play important roles in the model's
final performance, which motivates us to optimize them jointly.

On the other hand, the architecture of student-teacher pair is arbitrary
and fixed during training in conventional KD. However, recent works
\cite{cho2019efficacy,liu2020search} have pointed out the existence
of structural knowledge in KD, which implies that the teacher's architecture
has to match with the student to facilitate knowledge transfer. Therefore,
we are inspired to incorporate dynamic KD into our framework, where
the teacher is dynamically sampled to find the optimal matching for
the student.

We propose Joint-DetNAS, a unified framework consisting of two integrated
processes: \textbf{student morphism }and\textbf{ dynamic distillation}.\textbf{
Student morphism} aims to optimize the student's architecture while
remove the redundant parameters. To this end, an action space along
with a weight inheritance training strategy are carefully designed,
which eliminates the prerequisite of backbone's ImageNet pre-training
and allows the student to flexibly adjust its architecture while fully
utilize the predecessor's weights.\textbf{ Dynamic distillation} targets
at finding the optimal matching teacher and transferring its knowledge
to the student. To facilitate teacher search without repeated training,
an elastic teacher pool is built to provide sufficient powerful detectors,
which trains a super-network only once and obtains all the sub-networks
with competitive performances. During the search, we adopt a neat
hill climbing strategy to evolve the student-teacher pair. Thanks
to weight inheritance and the elastic teacher pool, each student-teacher
pair can be evaluated at the cost of fewer epochs and the final obtained
student detector requires no additional training

Our framework enables further exploration on the relationship between
the architectures of student-teacher pair. We observe two interesting
phenomena: (1) a more powerful detector does not necessarily make
a better teacher; (2) the capacities of the student and teacher are
highly correlated. These facts indicate the existence of structural
knowledge and architecture matching in KD for detection.

We conduct extensive experiments to verify the effectiveness of each
component (i.e., KD, pruning and the proposed elastic teacher pool)
on detection task. Our Joint-DetNAS presents clear performance enhancement
over 1) the input FPN baseline, 2) pipelining NAS-\textgreater pruning-\textgreater KD.
Given a classic R101-FPN as the base detector, our framework is able
to boost its AP from 41.4 to 43.9 on MS COCO and reduce its latency
by 47\%, which is on par with the SOTA EfficientDet \cite{Tan_2020_CVPR}
while requiring less search cost.

Our contributions are as follows: \textbf{1)} We investigate KD and
pruning for detection and carefully analyze their effectiveness. \textbf{2)}
We propose an elastic teacher pool containing sufficient powerful
detectors which can be directly sampled without training. \textbf{3)}
We develop a unified framework which jointly optimizes NAS, pruning
and dynamic KD. \textbf{4)} Extensive experiments are conducted to
investigate the matching pattern between the student-teacher pair
and verify the performance of our proposed framework.

\vspace{-1mm}

\section{Related Work}

\vspace{-1mm}

\begin{figure*}
\begin{centering}
\vspace{-2mm}
\includegraphics[scale=0.6]{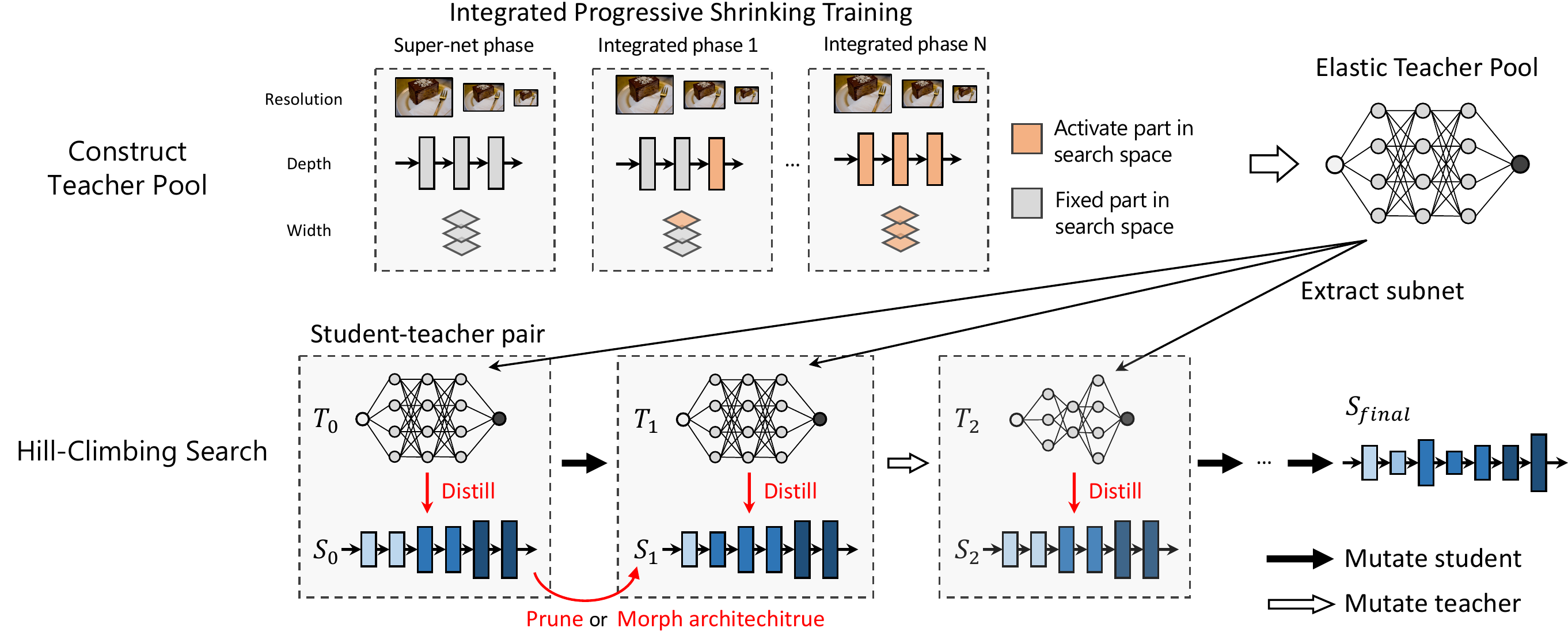}
\par\end{centering}
\vspace{-2mm}

\caption{\label{fig:framework}Illustration of our Joint-DetNAS. The algorithm
consists of student morphism and dynamic KD, which interleave with
each other. While student morphism optimizes the student's architecture,
dynamic distillation aims to find the optimal matching teacher. An
elastic teacher pool is trained via integrated progressive shrinking
strategy, from which teacher detectors can be sampled without additional
cost in subsequent searches; For student morphism, weight inheritance
strategy is adopted, allowing the student to flexibly update its architecture
while fully utilize the predecessor's weights.}

\vspace{-4mm}
\end{figure*}

\textbf{Object Detection.} State-of-the-art detection networks can
be classified as one-stage, two-stage and anchor-free detectors. One-stage
detectors such as \cite{redmon2017yolo9000,Liu2016,redmon2018yolov3}
directly makes prediction on the feature maps. Two-stage detectors
such as \cite{Ren2015,Lin2017a} uses a region proposal network (RPN)
to identify the foreground boxes and passes the corresponding features
to an RCNN head for final prediction. Recently, works such as \cite{tian2019fcos,duan2019centernet,yang2019reppoints}
propose to eliminate anchor priors and makes prediction directly.

\textbf{Neural Architecture Search. }NAS aims at finding an efficient
network architecture for a task automatically. There are numerous
works proposing different NAS methods for classification tasks \cite{baker2016designing,zoph2018learning,cai2018efficient,zhong2018practical,shi2019multi}
and detection tasks \cite{chen2018searching,liu2019auto,chen2019detnas}.
One recent paper \cite{liu2020search} proposed to combine NAS with
knowledge distillation by searching for the best student model given
a fixed teacher model, which also proves the existence of structural
knowledge in KD.

\textbf{Knowledge Distillation. }KD was first introduced in \cite{hinton2015distilling}
and its effectiveness for classification task has been validated by
extensive works \cite{zagoruyko2016paying,furlanello2018born,tarvainen2017mean,romero2014fitnets}.
However, few works have proposed KD methods for object detection \cite{chen2017learning,wang2019distilling,sun2020distilling},
which introduce only limited performance gain.

\textbf{Pruning. }Pruning methods have been well studied for classification
tasks \cite{molchanov2016pruning,li2016pruning,liu2017learning}.
which focus on reducing the model complexity without much performance
degradation. However, few works have verified its effectiveness on
detection tasks.

\vspace{-1mm}

\section{Proposed Method}

\subsection{The Joint-DetNAS framework}

\vspace{-1mm}

\subsubsection{Overview}

\vspace{-1mm}

As illustrated in Fig. \ref{fig:framework}, our Joint-DetNAS framework
comprises two core processes: \textbf{student morphism} and \textbf{dynamic
distillation}:

\textbf{Student morphism} aims to optimize the student's architecture
while reduce the redundant parameters. However, integrating the two
objectives is non-trivial: pruning requires pre-trained weights, which
is incompatible with current NAS paradigm, since it is practically
infeasible to obtain pre-trained weights that satisfy pruning requirements
for all sampled architectures. To address this issue, we propose a
carefully designed action space and a weight inheritance strategy,
which enable the student to flexibly adjust its architecture while
fully utilize the predecessor's weights. 

\textbf{Dynamic distillation} targets at finding the optimal matching
teacher to adapt to the student's structural changes, which calls
for a way of obtaining sufficient powerful teachers with low cost.
The mainstream NAS approach \cite{Ghiasi_2019_CVPR,yao2019sm,Tan_2020_CVPR}
using a proxy task (e.g., training with fewer epochs) to train the
teacher does not guarantee the quality of teacher's supervision. On
the other hand, training every teacher detector from scratch is too
costly. Therefore, inspired by the recent work \cite{cai2019once},
we propose to construct an elastic teacher pool (ETP) containing sub-networks
with high performances, which can be directly sampled as teachers
to supervise the student. Empowered by the proposed ETP, teachers
can be dynamically optimized according to the current status of the
students with high efficiency.

A neat hill climbing algorithm is adopted to integrate the two processes,
which enables adjusting the student's architecture and finding the
matching teacher simultaneously. Due to the use of weight inheritance
strategy and ETP, the search cost of our framework is significantly
reduced.

\vspace{-3mm}

\subsubsection{Student Morphism\label{subsec:Student-Morphism}}

Our goal is to adjust an input detector's backbone and enable better
adaptation to the given task. This is accomplished by continuously
applying beneficial actions to the backbone while fully utilize the
predecessor's parameters.

\begin{figure}
\begin{centering}
\vspace{-2mm}
\includegraphics[scale=0.65]{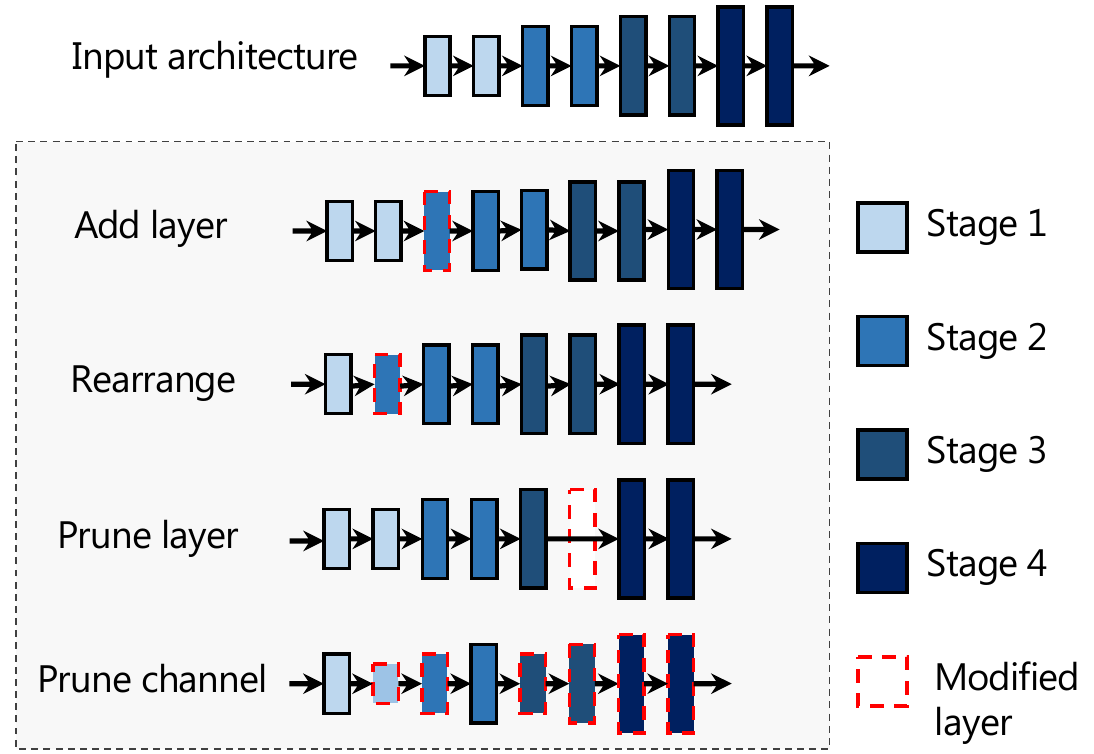}
\par\end{centering}
\vspace{-2mm}

\caption{\label{fig:student-action} Illustration of the student action space
$\mathbb{A}$. All actions are compatible with weight inheritance
strategy.}

\vspace{-4mm}
\end{figure}

\textbf{Action Space.} An action space $\mathbb{A}$ containing pruning
and network morphism is proposed to allow the student to flexibly
adjust its architecture while fully utilize the predecessor's weights.
\textbf{Pruning} removes the redundant parameters to make the model
compact\textbf{,} which includes 2 actions:\textbf{ (1) Layer Pruning}
directly removes a whole layer with least importance, while \textbf{(2)
Channel Pruning} removes the channels of convolutions which are insignificant.\textbf{
Network morphism }flexibly adjusts the student's architecture, two
actions are considered in this category: \textbf{(3)} \textbf{Add-Layer
}inserts a new layer at a given position and introduces more capacity
to enhance the model's performance; \textbf{(4) Rearrange }moves a
layer from one stage to its neighbor stage, which enables flexibly
re-allocating the backbone's computational budget for detection task.
The proposed action space supports various stage-based backbone families
for detector, e.g., ResNet, ResNeXt, and MobileNet series, etc.

\textbf{Weight Inheritance.} The trained weights of the predecessor
are inherited to (1) provide the initial pre-trained weights for pruning,
and (2) eliminate the expensive ImageNet pre-training prerequisites
for faster evaluation. Specifically, we define the inheritance process
as a function $f_{evolve}$: $f_{evolve}(S_{old}^{\text{\ensuremath{\theta}}},a)\longrightarrow$$S_{new}^{\theta^{'}}$,
which accepts a detector $S_{old}^{\text{\ensuremath{\theta}}}$ and
an action $a$ as its inputs and outputs a new detector $S_{new}^{\theta^{'}}$
with adjusted architecture and inherited parameters. The aforementioned
action space is highly compatible with Weight Inheritance and the
detail of $f_{evolve}$ for each action is elaborated in the appendix.

\textbf{Search with Dynamic Resolution. }The resolution of input images
play an important role in the performance and inference speed of detectors.
Instead of directly incorporating input resolutions into the search
process, which expands the search space considerably, we propose to
train the student by dynamically sampling a resolution in each training
iteration. Thus, multiple resolutions can be evaluated after training,
which boosts the search efficiency.

\vspace{-3mm}

\subsubsection{Dynamic distillation with Elastic Teacher Pool}

We are inspired by the recent work \cite{cai2019once}, in which a
progressive shrinking strategy is proposed to train a super-network
only once and obtain all the subnets with competitive performances.
This approach fits our requirement for building a pool of teachers
containing sufficient powerful detectors. However, the complexity
of detection task initiates new challenges for this already complicated
pipeline. 

\textbf{Subnet space. }For a backbone with multiple stages, each subnet
is determined by sampling the width and depth in a given range at
each stage, while the combination of all the subnets form the subnet
space. Other than the backbone, the FPN (neck) also plays an important
role in detection, which fuses the features maps of different scales
to obtain richer spatial information. Thus, we incorporate its widths
variations in our implementation. To facilitate the search, we design
our subnet space to cover architectures ranging from that of ResNet18
(1.0x width) to ResNet101 (1.5x width), which contains roughly 765000
networks (including different image resolutions) with competitive
performances. More implementation details of the subnet space can
be found in Section \ref{sec:Experiments} and the Appendix.

\textbf{Training with Integrated Progressive Shrinking (IPS). }The
training is divided into several phases: In the first phase, only
the largest super-net is trained; In the following phases, subnets
with shrunk depths and widths are gradually added into the subnet
space, while the super-net acts as the teacher to distill all subnets
using our KD method proposed in \ref{subsec:Knowledge-Distillation}.
In contrast to the progressive shrinking (PS) strategy proposed in
\cite{cai2019once}, where the shrinkage of width and depth are performed
sequentially, we propose an integrated progressive shrinking strategy
(IPS) to jointly optimize smaller depths and widths, thus significantly
reduces the training cost. More details can be found in the Appendix.

\vspace{-4mm}

\subsubsection{Search Algorithm}

\begin{algorithm}
\caption{\label{alg:algorithm-1}Hill Climbing Search of Joint-DetNAS}

\begin{algorithmic}[1] 

\State \textbf{Input}: base detector $S_{base}^{\theta}$; student
action space $\mathbb{A}$; resolution choices $R=\{r_{i}\}_{i=1,..,k}$;
an elastic teacher pool \textbf{$P$} with $P_{super}$ as the largest
super-net.

\State top-k-list $\leftarrow$ $\textrm{Ø}$; $\left\{ S_{old}^{\theta},T_{old}\right\} $
$\leftarrow$ $\left\{ S_{base}^{\theta},P_{super}\right\} $; 

\State Start hill climbing

\Repeat

\State $\left\{ S_{old}^{\theta},T_{old}\right\} $ $\leftarrow$
sample from top-$k$ list

\State choice $\leftarrow$ select to evolve teacher or student

\If {choice is student}

\State $a$ $\leftarrow$ sample from $\mathbb{A}$

\State $S_{new}^{\theta^{'}}$ $\leftarrow$ $f_{evolve}\left(S_{old}^{\theta},a\right)$;
$T_{new}$ $\leftarrow$ $T_{old}$

\Else

\State $T_{new}$ $\leftarrow$ mutate $T_{old}$ 

\State Extract $T_{new}$ from \textbf{$P$}; $S_{new}^{\theta^{'}}$
$\leftarrow$ $S_{old}^{\theta}$

\EndIf

\State Fast evaluate $\left\{ S_{new}^{\theta^{'}},T_{new}\right\} $
with all $r_{i}$

\State $s$$\leftarrow$ $\underset{i}{max}\left(H\left(S_{new}^{\theta^{'}},r_{i}\right)\right)$

\State Update top-$k$ list with $\left\{ S_{new}^{\theta^{'}},T_{new},s\right\} $

\Until Convergence

\end{algorithmic}
\end{algorithm}

\vspace{-1mm}

Different from mainstream NAS methods, our framework aims to upgrade
a base detector \textbf{$S_{base}$} rather than exploring the whole
search space. Comparing with sample-based search algorithms (e.g.,
RL \cite{zoph2016neural,tan2018mnasnet,Ghiasi_2019_CVPR}, BO \cite{shi2019bridging},
etc. ), the Hill Climbing (HL) approach efficiently evolves the student-teacher
pairs and is highly compatible with weight inheritance strategy.

During the search, we optimize the student and the teacher alternatively.
Specifically, the algorithm starts with an initial student-teacher
pair. During each iteration, either the student is updated by applying
an action (as described in Section \ref{subsec:Student-Morphism})
or the teacher is mutated by modifying the depth or width in each
backbone stage. Benefiting from the weight inheritance strategy, each
student-teacher pair can be evaluated with only a few epochs of training.
We use the following scoring metric to evaluate a student-teacher
pair:

\vspace{-5mm}

\[
H\left(S,R\right)=mAP\left(S\right)\times\left[\left(\frac{C\left(S\right)}{C_{base}}\right)\times\left(\frac{R}{R_{base}}\right)^{\beta}\right]^{-\alpha}
\]
where $S$ is the student detector; $C$ is the complexity metric,
which we adopt $FLOPS$ since we do not target any particular device;
$R$ is the resolution of input image; $C_{base}$ and $R_{base}$
are the base complexity and base resolution; $\alpha$ is a coefficient
that balances the performance and complexity trade-off; $\beta$ balances
the complexity introduced by the architecture and the input resolution.

The search procedure is illustrated in Algorithm \ref{alg:algorithm-1}.
Our framework can be parallelized on multiple machines to boost the
search efficiency.

\vspace{-1mm}

\subsection{Knowledge Distillation for Detection\label{subsec:Knowledge-Distillation}}

Detection KD requires delicate design to distill spatial and localization
information. Our detection KD method includes two components: a) \textbf{Feature-level
distillation} maximizes the agreement between teacher and student's
backbone features in interested areas; b) \textbf{Prediction-level
distillation} uses predictions outputs from teacher's heads as soft
labels to train the students.

\vspace{-4mm}

\subsubsection{Feature-level Distillation}

Feature maps encode important semantic information. However, imitating
the whole feature maps is hindered by severe imbalance between the
foreground instances and background regions. To this end, we only
distill the features of object proposals, the objective can be formulated
as:

\vspace{-5mm}

\[
L_{feat}=\frac{1}{N_{p}}\sum_{l=1}^{L}\sum_{i=1}^{W}\sum_{j=1}^{H}\sum_{c=1}^{C}\left(f_{adap}\left(F_{S}^{l}\right)_{ijc}-\left(F_{T}^{l}\right)_{ijc}\right)^{2}
\]
where $F_{S}$ and $F_{T}$ are features after ROI align; $f_{adap}(\cdot)$
is an adaptation function mapping $F_{S}$ and $F_{T}$ to the same
dimension; $N_{p}$ is the number of mask's positive points; $L$
is number of FPN layers; $W,H,C$ are feature dimensions.\vspace{-4mm}

\subsubsection{Prediction-level Distillation}

The prediction level KD loss can be expressed in terms of classification
and regression KD loss: $L_{pred}=L_{cls}+L_{loc}$.

\textbf{Uncertainty from Classification}. Similar to classification,
the student is optimized by soft cross entropy loss using teacher's
logits as targets, which can be written as: $L_{cls}=-\frac{1}{N}\sum_{i}^{N}\mathbf{P}_{t}^{i}\log\mathbf{P}_{s}^{i}$,
where $N$ is the number of training data; $\mathbf{P}_{t}$ and $\mathbf{P}_{s}$
are predicted score vectors of the teacher and the student, respectively.

\textbf{Uncertainty from Localization}. Simply imitating the four
coordinates from teacher's outputs provides limited information about
how teacher localize objects, which motivates us to incorporate the
class ``uncertainty'' knowledge into this process, i.e., utilizing
prediction for all classes generated by the regression decoder. The
class-aware localization outputs encode the teacher's ability of localizing
proposals (can be viewed as a parts of objects) given different class
hypotheses. Specifically, we calculate the sum of regression values
weighted by classification scores: $L_{reg}=\frac{1}{N}\sum^{N}\mid\sum_{i=0}^{C}p_{t}^{i}\times\left(reg_{t}^{i}-reg_{s}^{i}\right)\mid$,
where $C$ is the number of classes; $p_{i}$ and $reg_{i}$ are the
classification score and regression outputs of foreground class $i$;
superscripts $s$ and $t$ stand for student and teacher.

\vspace{-1mm}

\subsection{Model Pruning}

Pruning is incorporated into the framework as a part of student morphism
to reduce student detector's complexity. We utilize both layer-wise
and channel-wise pruning, which reduce the student's depth and width
respectively. \textbf{Layer-wise Pruning }removes backbone's entire
layer with the least L1-Norm.\textbf{ Channel-wise Pruning }reduces
the width of detector's backbone, for which we apply network slimming
approach \cite{liu2017learning}. The method determines the channel
importance according to the magnitude of BN's weights. Then the channels
with least importance are removed. To encourage channel sparsity,
we add a regularization loss to BN's weight parameters $\gamma$:
$L_{BN}=\sum_{\gamma\in\Gamma}\mid\gamma\mid$. A small pruning percentage
is set during each student morphism to progressively shrink the student
without causing much performance deterioration.

\vspace{-4mm}

\paragraph{Overall Loss for Training Students}

The total loss for training student detectors can be represented as:
$L=L_{det}+L_{feat}+L_{pred}+\lambda L_{BN}$, where $L_{det}$ denotes
the normal detection training loss; $\lambda$ is the coefficient
of the regularization loss for pruning, which is set to 0.00001. $L$
is enforced on the student throughout the search process.

\begin{table}
\vspace{-2mm}

\begin{centering}
\tabcolsep 0.03in{\footnotesize{}}%
\begin{tabular}{c|c|c|c}
\hline 
{\small{}Base model} & {\small{}Method} & {\small{}Input size} & {\small{}${\rm AP}$}\tabularnewline
\hline 
\hline 
\multirow{2}{*}{{\small{}R18-FPN}} & {\small{}standard} & {\footnotesize{}$800\times600/1333\times800$} & {\small{}34.3/36.0}\tabularnewline
 & {\small{}our ETP} & {\footnotesize{}$800\times600/1333\times800$} & {\small{}35.1$^{+0.8}$/36.1$^{+0.1}$}\tabularnewline
\hline 
\multirow{2}{*}{{\small{}R50-FPN}} & {\small{}standard} & {\footnotesize{}$800\times600/1333\times800$} & {\small{}38.4/39.5}\tabularnewline
 & {\small{}our ETP} & {\footnotesize{}$800\times600/1333\times800$} & {\small{}41.8$^{+3.4}$/42.4$^{+2.9}$}\tabularnewline
\hline 
\multirow{2}{*}{{\small{}R101-FPN}} & {\small{}standard} & {\footnotesize{}$800\times600/1333\times800$} & {\small{}39.7/41.4}\tabularnewline
 & {\small{}our ETP} & {\footnotesize{}$800\times600/1333\times800$} & {\small{}43.1$^{+3.4}$/44.1$^{+2.7}$}\tabularnewline
\hline 
\end{tabular}{\footnotesize\par}
\par\end{centering}
\vspace{-2mm}

\caption{\label{tab:ofa}The subnets sampled from elastic teacher pool (ETP)
consistently outperform their equivalent baselines trained under standard
training strategy (2x+ms).}
\vspace{-2mm}
\end{table}

\begin{table}
\begin{centering}
\tabcolsep 0.02in{\scriptsize{}}%
\begin{tabular}{c|c|c|c}
\hline 
\multirow{2}{*}{{\small{}Model}} & {\small{}Pruning} & {\small{}Backbone} & \multirow{2}{*}{{\small{}${\rm AP}$}}\tabularnewline
 & {\small{}percentage} & {\small{}FLOPS(G)} & \tabularnewline
\hline 
\multirow{2}{*}{{\small{}R50-FPN}} & {\small{}0\% (baseline)} & {\small{}84.1} & {\small{}37.1}\tabularnewline
 & {\small{}10\%/20\%/30\%} & {\small{}74.4/70.0/65.8} & {\small{}37.3/37.0/36.4}\tabularnewline
\hline 
\multirow{2}{*}{{\small{}R101-FPN}} & {\small{}0\% (baseline)} & {\small{}160.2} & {\small{}39.0}\tabularnewline
 & {\small{}10\%/20\%/30\%} & {\small{}140.9/125.7/112.5} & {\small{}39.2/38.8/38.3}\tabularnewline
\hline 
\end{tabular}{\scriptsize\par}
\par\end{centering}
\vspace{-2mm}

\caption{\label{tab:pruning}Pruning results for R50-FPN and R101-FPN given
different channel pruning percentages. The detector's FLOPS can be
effectively reduced without much performance degradation.}
\vspace{-2mm}
\end{table}

\begin{table}
\begin{centering}
\tabcolsep 0.1in{\scriptsize{}}%
\begin{tabular}{c|c|c|c}
\hline 
{\small{}KD Method} & {\small{}Student} & {\small{}Teacher} & {\small{}${\rm AP}$}\tabularnewline
\hline 
\hline 
{\small{}FGFI \cite{wang2019distilling}} & {\small{}R50-half} & {\small{}R50} & {\small{}34.8}\tabularnewline
{\small{}TAR \cite{sun2020distilling}} & {\small{}R50} & {\small{}R152+R101} & {\small{}40.1}\tabularnewline
\hline 
\multirow{2}{*}{\textbf{\small{}our KD}} & {\small{}R18 (36.0)} & {\small{}R50 (39.5)} & \textbf{\small{}38.1$^{+2.1}$}\tabularnewline
 & {\small{}R50 (39.5)} & {\small{}R101 (41.4)} & \textbf{\small{}41.6$^{+2.1}$}\tabularnewline
\hline 
\end{tabular}{\scriptsize\par}
\par\end{centering}
\vspace{-2mm}

\caption{\label{tab:distillation-1}Comparison between our detection KD method
with baselines and previous KD works. The values in parentheses are
the baseline AP and teacher's AP, respectively. Our KD method outperforms
others by a large margin.}
\vspace{-2mm}
\end{table}

\vspace{-2mm}

\section{Experiments\label{sec:Experiments}}

\textbf{Datasets and evaluation metrics.} We use MS COCO \cite{Lin2014}
to conduct experiments. The mAP for IoU thresholds from 0.5 to 0.95
is used as the performance metric.

\textbf{Implementation details}. We use ResNet-based detectors to
construct our elastic teacher pool, the subnet space for the backbone
contains depth ranging from {[}2,2,2,2{]} to {[}3,4,23,3{]} for four
stages, and the width for each backbone stage and the neck can be
sampled from {[}$W$, $1.25\times W$, $1.5W${]}, where $W$ is the
width of standard ResNet. During search, each student-teacher pair
is trained with 3 epochs for fast evaluation. For each teacher subnet
sampled from the teacher pool, we reset its BN statistics by forwarding
a batch of images, which is essential for performance recovery. More
details are in the Appendix.

\vspace{-2mm}

\subsection{Ablation Study}

\begin{figure}
\begin{centering}
\vspace{-2mm}
\par\end{centering}
\begin{centering}
\includegraphics[scale=0.28]{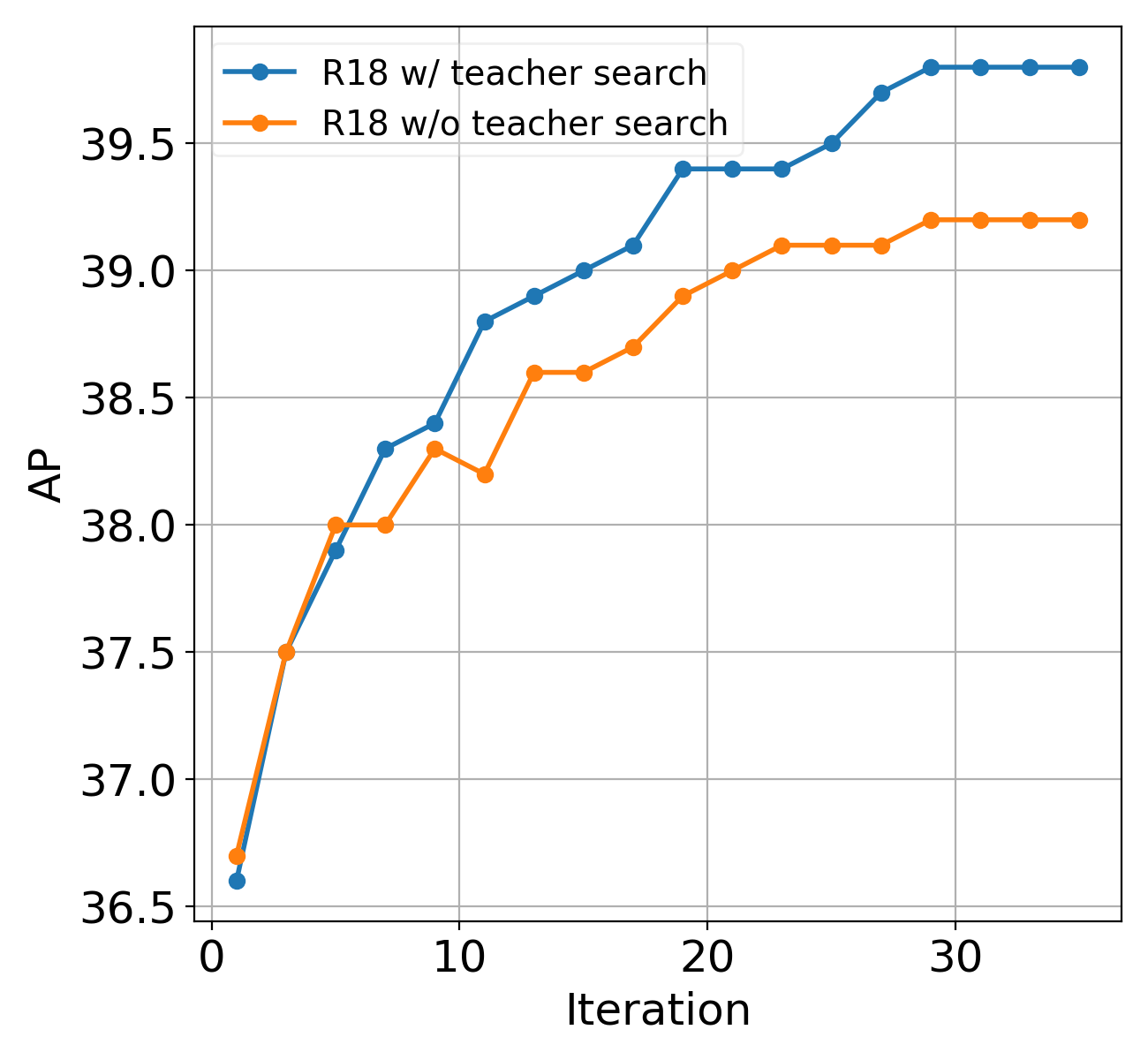}
\par\end{centering}
\begin{centering}
\vspace{-2mm}
\par\end{centering}
\caption{\label{fig:teacher-search-benefit}The comparison between dynamic
KD and convention KD (with the super-net of ETP as the teacher). Dynamic
KD can boost the student faster and help it reach a higher final performance.}

\begin{centering}
\vspace{-3mm}
\par\end{centering}
\end{figure}

\subsubsection{Decoupling the Framework}

Each component plays an important role in the overall Joint-DetNAS
framework. Thus, it is essential to decouple them from the framework
and separately analyze their effectiveness in detail.

\textbf{Quality of Elastic Teacher Pool. }Our framework requires the
teacher detectors sampled from the ETP to have competitive performances.
To demonstrate the quality of our ETP, we compare its sampled subnets
with their equivalent classic FPN detectors trained under standard
2x schedule and multi-scale training (for easier notation, we denote
this as 2x+ms in later sections) strategy in Table \ref{tab:ofa}.
The former consistently outperforms the latter.

\textbf{Pruning.} We conduct experiments to prune the backbone of
R50-FPN and R101-FPN detectors given different channel pruning percentages
in Table \ref{tab:pruning}. The detectors are pre-trained for 12
epochs before pruning and fine-tuned for extra 3 epochs afterwards.
The detector's parameter can be effectively reduced without much performance
degradation. e.g., For both detectors, the performance after pruning
30\% channels is still comparable to the original.

\textbf{Distillation.} Our detection KD framework is simple yet effective.
We compare our detection KD method with baselines and previous KD
works in Table \ref{tab:distillation-1}. The R18-FPN and R50-FPN
detectors are adopted as the students, with R50-FPN and R101-FPN as
the teachers, respectively. To demonstrate effectiveness of our KD
method, stronger baselines (2x+ms) are used. The results show that
our method outperform the others by a large margin.

\vspace{-2mm}

\subsubsection{Dynamic KD Benefits the Student\label{subsec:Dynamic-KD-Benefits}}

We aim to verify the superiority of dynamic KD: whether dynamic teacher
is better than a fixed powerful teacher for transferring knowledge.
Specifically, we fix the ResNet18-FPN detector as the student and
follow the 3-epoch iterative training schedule, then conduct two experiments
(1) dynamic KD (DKD): the teacher is dynamically sampled in every
iteration and (2) Conventional KD (CKD): the largest super-net in
ETP is used as the teacher. The results in Figure \ref{fig:teacher-search-benefit}
shows that DKD can boost the student faster and help it reach a higher
final performance. This also implies the underlying structural knowledge
in KD, for which we provide further analysis in later Section \ref{subsec:Teacher-student-relationship}.

\subsection{Main Results}

\subsubsection{Comparison with Baselines}

Joint-DetNAS can upgrade detectors with various backbone designs.
We conduct experiments on FPN detectors with R18, R50, R101 and X101
as backbones to verify the effectiveness of our framework. We use
$1333\times800$ resolution with 2x+ms training for baseline and compare
with our result using searched resolution. As shown in Table \ref{tab:compare-baseline},
our method consistently boosts the detectors' performances while substantially
reduces their complexities. Notably, for R101-FPN, the upgraded detector
achieves $+2.5$ gain in $AP$ and $47\%$ reduction in latency.

\begin{table}
\vspace{-2mm}

\begin{centering}
\tabcolsep 0.03in{\scriptsize{}}%
\begin{tabular}{c|c|c|c|c|c}
\hline 
{\footnotesize{}Base model} & {\footnotesize{}Group} & {\footnotesize{}Input size} & {\footnotesize{}FLOPS (G)} & {\footnotesize{}FPS} & {\footnotesize{}${\rm AP}$}\tabularnewline
\hline 
\hline 
\multirow{2}{*}{{\footnotesize{}R18-FPN}} & {\footnotesize{}baseline} & {\footnotesize{}$1333\times800$} & {\footnotesize{}160.5} & {\footnotesize{}28.2} & {\footnotesize{}36.0}\tabularnewline
 & \textbf{\footnotesize{}ours} & {\footnotesize{}$1080\times720$} & \textbf{\footnotesize{}117.3$^{-27\%}$} & \textbf{\footnotesize{}33.0$^{+17\%}$} & \textbf{\footnotesize{}38.5$^{+2.5}$}\tabularnewline
\hline 
\multirow{2}{*}{{\footnotesize{}R50-FPN}} & {\footnotesize{}baseline} & {\footnotesize{}$1333\times800$} & {\footnotesize{}215.8} & {\footnotesize{}20.5} & {\footnotesize{}39.5}\tabularnewline
 & \textbf{\footnotesize{}ours} & {\footnotesize{}$1080\times720$} & \textbf{\footnotesize{}145.7$^{-32\%}$} & \textbf{\footnotesize{}25.4$^{+24\%}$} & \textbf{\footnotesize{}42.3$^{+2.8}$}\tabularnewline
\hline 
\multirow{2}{*}{{\footnotesize{}R101-FPN}} & {\footnotesize{}baseline} & {\footnotesize{}$1333\times800$} & {\footnotesize{}295.7} & {\footnotesize{}15.9} & {\footnotesize{}41.4}\tabularnewline
 & \textbf{\footnotesize{}ours} & {\footnotesize{}$1080\times720$} & \textbf{\footnotesize{}153.9$^{-48\%}$} & \textbf{\footnotesize{}23.3$^{+47\%}$} & \textbf{\footnotesize{}43.9$^{+2.5}$}\tabularnewline
\hline 
\multirow{2}{*}{{\footnotesize{}X101-FPN}} & {\footnotesize{}baseline} & {\footnotesize{}$1333\times800$} & {\footnotesize{}286.9} & {\footnotesize{}13.2} & {\footnotesize{}42.9}\tabularnewline
 & \textbf{\footnotesize{}ours} & {\footnotesize{}$1333\times800$} & \textbf{\footnotesize{}266.3$^{-7\%}$} & \textbf{\footnotesize{}14.0$^{+6\%}$} & \textbf{\footnotesize{}45.7$^{+2.8}$}\tabularnewline
\hline 
\end{tabular}{\scriptsize\par}
\par\end{centering}
\vspace{-1mm}

\caption{\label{tab:compare-baseline} Our Joint-DetNAS can upgrade detectors
with various backbone designs. Joint-DetNAS consistently boosts the
input baseline detectors' performances as well as substantially reduces
their complexities. }
\vspace{-2mm}
\end{table}

\vspace{-2mm}

\subsubsection{Joint Optimization Beats Naive Pipelining}

Intuitively, NAS, pruning and KD is can be combined by pipelining:
first search a detector with NAS, then prune it and train it with
KD. We compare our joint optimization approach with pipelining methods:
(1) Start with regular R101-FPN detector or a NAS-searched detector
with lower complexity; (2) pre-train them with the pruning regularization
loss; (3) prune the detector to comparable complexity with the result
of Joint-DetNAS (R101-based); (4) train the pruned detector with the
proposed KD under standard training strategy (2x+ms) and the same
resolution ($1080\times720$). In Table \ref{tab:compare-baseline},
we compare the result of NAS-prune-KD and R101-prune-KD and find that
the performance gain brought by NAS diminishes after pruning and KD
are applied, indicating that the naive pipelining strategy leads to
suboptimal. In contrast, our joint optimization methods outperforms
both pipelining methods by a large margin.

\begin{table}
\vspace{-2mm}

\begin{centering}
\tabcolsep 0.02in{\scriptsize{}}%
\begin{tabular}{c|c|c|c|c}
\hline 
\multirow{4}{*}{{\small{}Method}} & \multicolumn{2}{c|}{{\small{}Intermediate }} & \multicolumn{2}{c}{{\small{}Final }}\tabularnewline
 & \multicolumn{2}{c|}{{\small{}(pre-training)}} & \multicolumn{2}{c}{{\small{}(w/ prune+KD)}}\tabularnewline
\cline{2-5} \cline{3-5} \cline{4-5} \cline{5-5} 
 & {\small{}backbone} & \multirow{2}{*}{{\small{}AP}} & {\small{}backbone} & \multirow{2}{*}{{\small{}AP}}\tabularnewline
 & {\small{}FLOPS (G)} &  & {\small{}FLOPS (G)} & \tabularnewline
\hline 
\hline 
{\small{}R101-prune-KD} & {\small{}122.6} & {\small{}39.0} & {\small{}60.8$^{-50\%}$} & {\small{}42.1$^{+3.1}$}\tabularnewline
{\small{}NAS-prune-KD} & {\small{}105.2} & {\small{}39.9} & {\small{}59.1$^{-44\%}$} & {\small{}41.8$^{+1.9}$}\tabularnewline
\multirow{1}{*}{\textbf{\small{}Joint-DetNAS (R101)}} & {\small{}-} & {\small{}-} & \textbf{\small{}58.9} & \textbf{\small{}43.9}\tabularnewline
\hline 
\end{tabular}{\scriptsize\par}
\par\end{centering}
\vspace{-1mm}

\caption{\label{tab:compare-baseline-1}Comparison between Joint-DetNAS and
the naive pipelining approach. The results show that the pipelining
methods leads to suboptimal, while Joint-DetNAS is capable of better
integrating NAS, pruning and KD.}
\vspace{-2mm}
\end{table}

\vspace{-3mm}

\subsubsection{Comparison with State-of-the-art}

We compare our method with the SOTA manually designed detectors (e.g.,
FCOS\cite{tian2019fcos}, RepPoints\cite{yang2019reppoints} and CB-Net\cite{liu2020cbnet},
etc.) and NAS-based (e.g., NAS-FPN\cite{Ghiasi_2019_CVPR}, SP-NAS\cite{jiang2020sp},
etc.) approaches. The results of the COCO's \texttt{test-dev} split
are reported in Table \ref{tab:compare-sota}. Our Joint-DetNAS outperforms
SOTA manually designed detectors in terms of both FPS and AP, e.g.
our searched detector based on R101 reaches 23.3 FPS and 43.9 AP,
outperforming RepPoint-R101's 13.7 FPS and 41.0 AP by a large margin.
Furthermore, our method (R101-based) surpasses most mainstream detection
NAS methods (e.g., SM-NAS \cite{yao2019sm} and NAS-FPN \cite{Ghiasi_2019_CVPR})
and reaches comparable performance with the SOTA EfficientDet (D2)
\cite{Tan_2020_CVPR}, while requiring much less search cost and no
extra post-search training epochs.

\begin{table*}
\vspace{-2mm}

\begin{centering}
\tabcolsep 0.03in%
\begin{tabular}{c|c|c|c|c|c|c|c|c|c|c}
\hline 
\multirow{3}{*}{{\footnotesize{}Method}} & \multirow{3}{*}{{\footnotesize{}Backbone}} & \multirow{3}{*}{{\footnotesize{}Input size}} & {\footnotesize{}Post-search} & \multirow{3}{*}{{\footnotesize{}FPS}} & \multirow{3}{*}{{\footnotesize{}${\rm AP}$}} & \multirow{3}{*}{{\footnotesize{}${\rm AP_{@.5}}$}} & \multirow{3}{*}{{\footnotesize{}${\rm AP_{@.7}}$}} & \multirow{3}{*}{{\footnotesize{}${\rm AP_{S}}$}} & \multirow{3}{*}{{\footnotesize{}${\rm AP_{M}}$}} & \multirow{3}{*}{{\footnotesize{}${\rm AP_{L}}$}}\tabularnewline
 &  &  & {\footnotesize{}training } &  &  &  &  &  &  & \tabularnewline
 &  &  & {\footnotesize{}epochs} &  &  &  &  &  &  & \tabularnewline
\hline 
\multicolumn{11}{c}{\textbf{\footnotesize{}Manually Designed}}\tabularnewline
\hline 
{\footnotesize{}Cascade RCNN \cite{cai2017cascade}} & {\footnotesize{}R101} & {\footnotesize{}$1333\times800$} & {\footnotesize{}-} & {\footnotesize{}13.5 (V100)$^{\dagger}$} & {\footnotesize{}43.6} & {\footnotesize{}62.1} & {\footnotesize{}47.4} & {\footnotesize{}26.1} & {\footnotesize{}47.0} & {\footnotesize{}53.6}\tabularnewline
{\footnotesize{}FCOS \cite{tian2019fcos}} & {\footnotesize{}R101} & {\footnotesize{}$1333\times800$} & {\footnotesize{}-} & {\footnotesize{}17.3 (V100)$^{\dagger}$} & {\footnotesize{}41.5} & {\footnotesize{}60.7} & {\footnotesize{}45.0} & {\footnotesize{}24.4} & {\footnotesize{}44.8} & {\footnotesize{}51.6}\tabularnewline
{\footnotesize{}RepPoints \cite{yang2019reppoints}} & {\footnotesize{}R101} & {\footnotesize{}$1333\times800$} & {\footnotesize{}-} & {\footnotesize{}13.7 (V100)$^{\dagger}$} & {\footnotesize{}41.0} & {\footnotesize{}62.9} & {\footnotesize{}44.3} & {\footnotesize{}23.6} & {\footnotesize{}44.1} & {\footnotesize{}51.7}\tabularnewline
{\footnotesize{}CB-Net w/ Cascade \cite{liu2020cbnet}} & {\footnotesize{}R101-TB} & {\footnotesize{}$1333\times800$} & {\footnotesize{}-} & {\footnotesize{}5.5 (V100)$^{\ddagger}$} & {\footnotesize{}44.9} & {\footnotesize{}63.9} & {\footnotesize{}48.9} & {\footnotesize{}-} & {\footnotesize{}-} & {\footnotesize{}-}\tabularnewline
\hline 
\multicolumn{11}{c}{\textbf{\footnotesize{}NAS-Based}}\tabularnewline
\hline 
{\footnotesize{}Det-NAS \cite{chen2019detnas}} & {\footnotesize{}DetNASNet} & {\footnotesize{}$1333\times800$} & {\footnotesize{}24} & {\footnotesize{}20.4 (V100)$^{\ddagger}$} & {\footnotesize{}40.2} & {\footnotesize{}61.5} & {\footnotesize{}43.6} & {\footnotesize{}23.3} & {\footnotesize{}42.5} & {\footnotesize{}53.8}\tabularnewline
{\footnotesize{}SM-NAS (E5) \cite{yao2019sm}} & {\footnotesize{}SMNet (searched)} & {\footnotesize{}$1333\times800$} & {\footnotesize{}24} & {\footnotesize{}9.3 (V100)$^{\ddagger}$} & {\footnotesize{}45.9} & {\footnotesize{}64.6} & {\footnotesize{}49.6} & {\footnotesize{}27.1} & {\footnotesize{}49.0} & {\footnotesize{}58.0}\tabularnewline
{\footnotesize{}SP-NAS \cite{jiang2020sp}} & {\footnotesize{}SPNet-XB} & {\footnotesize{}$1333\times800$} & {\footnotesize{}24} & {\footnotesize{}5.6 (V100)$^{\ddagger}$} & {\footnotesize{}47.4} & {\footnotesize{}65.7} & {\footnotesize{}51.9} & {\footnotesize{}29.6} & {\footnotesize{}51.0} & {\footnotesize{}60.4}\tabularnewline
{\footnotesize{}NAS-FPN (7@384) \cite{Ghiasi_2019_CVPR}} & {\footnotesize{}AmoebaNet} & {\footnotesize{}$1280\times1280$} & {\footnotesize{}150} & {\footnotesize{}3.6 (P100)$^{\ddagger}$} & {\footnotesize{}48.0} & {\footnotesize{}-} & {\footnotesize{}-} & {\footnotesize{}-} & {\footnotesize{}-} & {\footnotesize{}-}\tabularnewline
{\footnotesize{}EfficientDet (D2)$^{*}$ \cite{Tan_2020_CVPR}} & {\footnotesize{}EfficientNet (B2)} & {\footnotesize{}$768\times768$} & {\footnotesize{}300} & {\footnotesize{}26.8 (V100)$^{\dagger}$}\tablefootnote{{\footnotesize{}The FPS of EfficientDet's Pytorch implementation \url{https://github.com/zylo117/Yet-Another-EfficientDet-Pytorch}
is reported for fair comparison.}} & {\footnotesize{}43.9} & {\footnotesize{}62.7} & {\footnotesize{}47.6} & {\footnotesize{}-} & {\footnotesize{}-} & {\footnotesize{}-}\tabularnewline
\hline 
\textbf{\footnotesize{}Joint-DetNAS (R50)} & {\footnotesize{}R50-searched} & \textbf{\footnotesize{}$1080\times720$} & {\footnotesize{}-} & \textbf{\footnotesize{}25.4}{\footnotesize{} (V100)$^{\dagger}$} & \textbf{\footnotesize{}42.3} & \textbf{\footnotesize{}62.6} & \textbf{\footnotesize{}46.2} & \textbf{\footnotesize{}26.2} & \textbf{\footnotesize{}45.1} & \textbf{\footnotesize{}50.6}\tabularnewline
\textbf{\footnotesize{}Joint-DetNAS (R101)} & {\footnotesize{}R101-searched} & \textbf{\footnotesize{}$1080\times720$} & {\footnotesize{}-} & \textbf{\footnotesize{}23.3}{\footnotesize{} (V100)$^{\dagger}$} & \textbf{\footnotesize{}43.9} & \textbf{\footnotesize{}63.8} & \textbf{\footnotesize{}47.9} & \textbf{\footnotesize{}27.0} & \textbf{\footnotesize{}46.8} & \textbf{\footnotesize{}52.8}\tabularnewline
\textbf{\footnotesize{}Joint-DetNAS (X101-Cascade)} & {\footnotesize{}X101-searched-DCN} & \textbf{\footnotesize{}$1333\times800$} & {\footnotesize{}16} & \textbf{\footnotesize{}10.1}{\footnotesize{} (V100)$^{\dagger}$} & \textbf{\footnotesize{}50.7} & \textbf{\footnotesize{}69.6} & \textbf{\footnotesize{}55.4} & \textbf{\footnotesize{}31.3} & \textbf{\footnotesize{}53.8} & \textbf{\footnotesize{}64.0}\tabularnewline
\hline 
\end{tabular}
\par\end{centering}
\vspace{-1mm}

\caption{\label{tab:compare-sota} Comparison with SOTA manually designed and
NAS-based methods. We obtain the X101-Cascade model by upgrading the
searched X101-based detector with DCN and Cascade head, and further
fine-tune it for 16 epochs with HTC \cite{chen2019hybrid} teacher.
FPS is reported with batch size 1; $^{\dagger}$ and $^{\ddagger}$
represent the results obtained on our own V100 device and from the
original paper, respectively. $^{*}$ means soft-NMS is adopted. }

\vspace{-2mm}
\end{table*}

\vspace{-2mm}

\subsubsection{Search Efficiency}

Search efficiency is a key issue in NAS. We compare Joint-DetNAS with
other SOTA detection NAS methods (e.g., \cite{chen2019detnas,yao2019sm})
in Table \ref{tab:compare-differen-search}. Our framework finds better
performance-complexity tradeoff for the detector with less search
cost,

\begin{table}
\vspace{-2mm}

\begin{centering}
\tabcolsep 0.01in{\scriptsize{}}%
\begin{tabular}{c|c|c|c|c}
\hline 
\multirow{2}{*}{{\footnotesize{}Search Method}} & \multirow{2}{*}{{\footnotesize{}FLOPS}} & \multirow{2}{*}{{\footnotesize{}${\rm AP}$}} & {\footnotesize{}\#Searched} & {\footnotesize{}Search cost}\tabularnewline
 &  &  & \multirow{1}{*}{{\footnotesize{}architectures}} & {\footnotesize{}(GPU days)}\tabularnewline
\hline 
{\footnotesize{}random} & {\footnotesize{}-} & {\footnotesize{}-} & {\footnotesize{}50} & {\footnotesize{}\textasciitilde 1200}\tabularnewline
{\footnotesize{}Det-NAS \cite{chen2019detnas}} & {\footnotesize{}289.4} & {\footnotesize{}40.0} & {\footnotesize{}1000} & {\footnotesize{}70}\tabularnewline
{\footnotesize{}NAS-FPN (R50-7@256) \cite{Ghiasi_2019_CVPR}} & {\footnotesize{}281.3} & {\footnotesize{}39.9} & {\footnotesize{}10000} & {\footnotesize{}\textgreater\textgreater 500}\tabularnewline
{\footnotesize{}SP-NAS \cite{jiang2020sp}} & {\footnotesize{}349.3} & {\footnotesize{}41.7} & {\footnotesize{}200} & {\footnotesize{}200}\tabularnewline
\textbf{\footnotesize{}Joint-DetNAS}{\footnotesize{} }\textbf{\footnotesize{}(R101-based)} & \textbf{\footnotesize{}145.7} & \textbf{\footnotesize{}43.9} & {\footnotesize{}100} & {\footnotesize{}200}\tabularnewline
\hline 
\end{tabular}
\par\end{centering}
\vspace{-1mm}

\caption{\label{tab:compare-differen-search} Comparison with other search
methods. The search cost consists of 3 parts: (1) pre-training cost
(including ImageNet pre-training or ETP training), NAS cost and post-training
cost. We only estimate the cost for random search as it is prohibitively
expensive (ImageNet pre-training for each sampled detector).}
\vspace{-2mm}
\end{table}

\subsection{Looking into the Search Results: More analysis}

\subsubsection{Teacher-student Relationship\label{subsec:Teacher-student-relationship}}

\begin{figure}
\begin{centering}
\vspace{-2mm}
\includegraphics[scale=0.27]{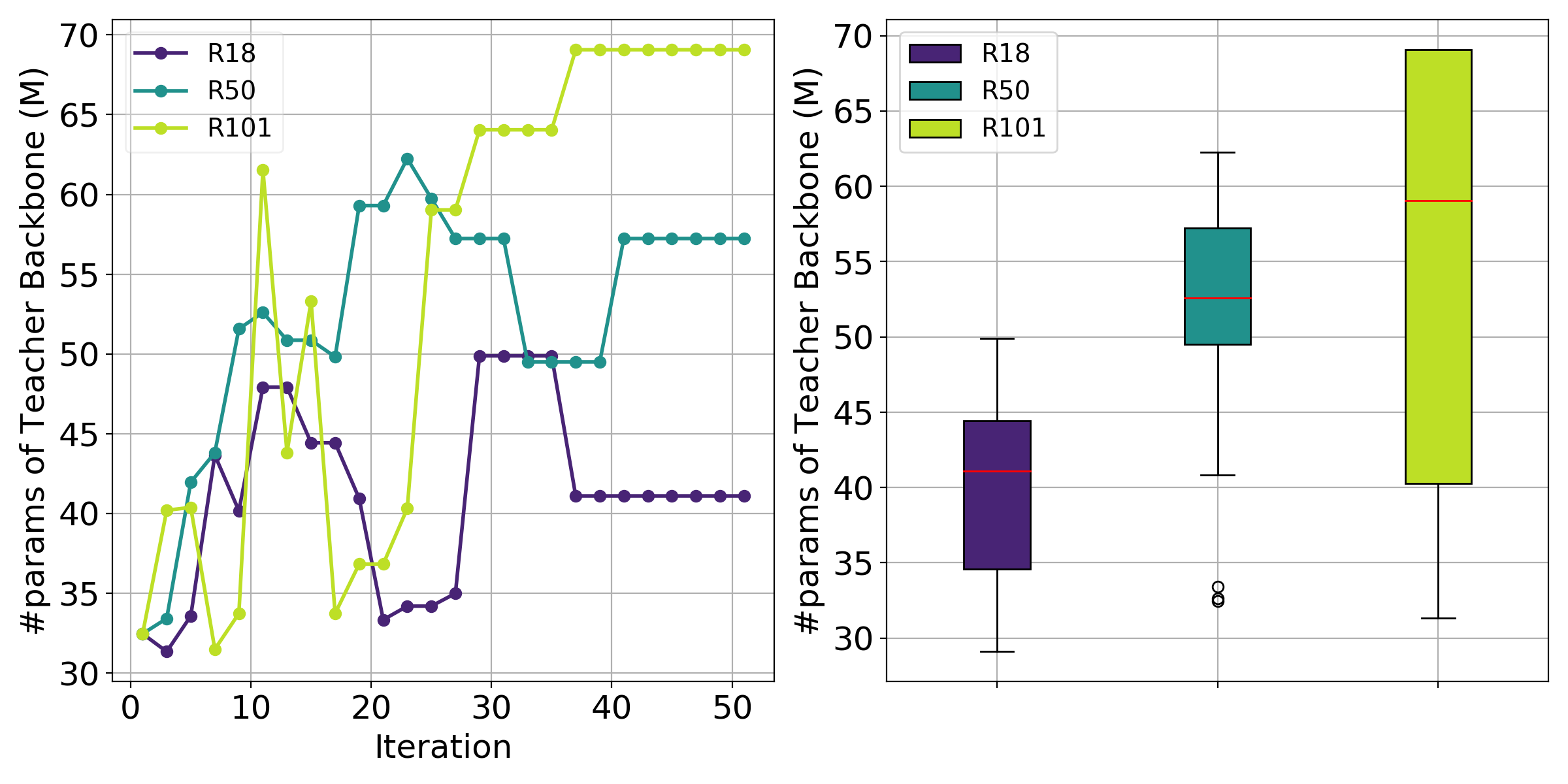}
\par\end{centering}
\vspace{-2mm}

\caption{\label{fig:student-teacher-relationship}The teacher-student capacity-matching
pattern during search. \textbf{Left}: the teacher's backbone parameters
of the best student in the current iteration. \textbf{Right}: distribution
of teacher's backbone parameters throughout the search. The matching
teacher's complexity is highly correlated with that of the student.}

\vspace{-3mm}
\end{figure}
As observed in earlier Section \ref{subsec:Dynamic-KD-Benefits},
larger detectors may not be better teachers, which naturally prompts
us to further explore the matching pattern of promising teachers for
different students. To this end, we apply dynamic KD to search optimal
matching teachers for students with various complexities (i.e., FPN
with R18, R50, R101 as the backbones). As shown in Figure \ref{fig:student-teacher-relationship},
starting with the same teacher, each student can converge to different
teachers. The results present a clear pattern: smaller students tend
to match teachers with lower capacities, and vice versa. This phenomenon
implies the underlying interdependence of complexity between the student-teacher
pairs, which can provide useful insights for designing detection KD
system.

\vspace{-2mm}

\subsubsection{How Students Evolve: Action Analysis}

\begin{figure}
\begin{centering}
\vspace{-2mm}
\includegraphics[scale=0.23]{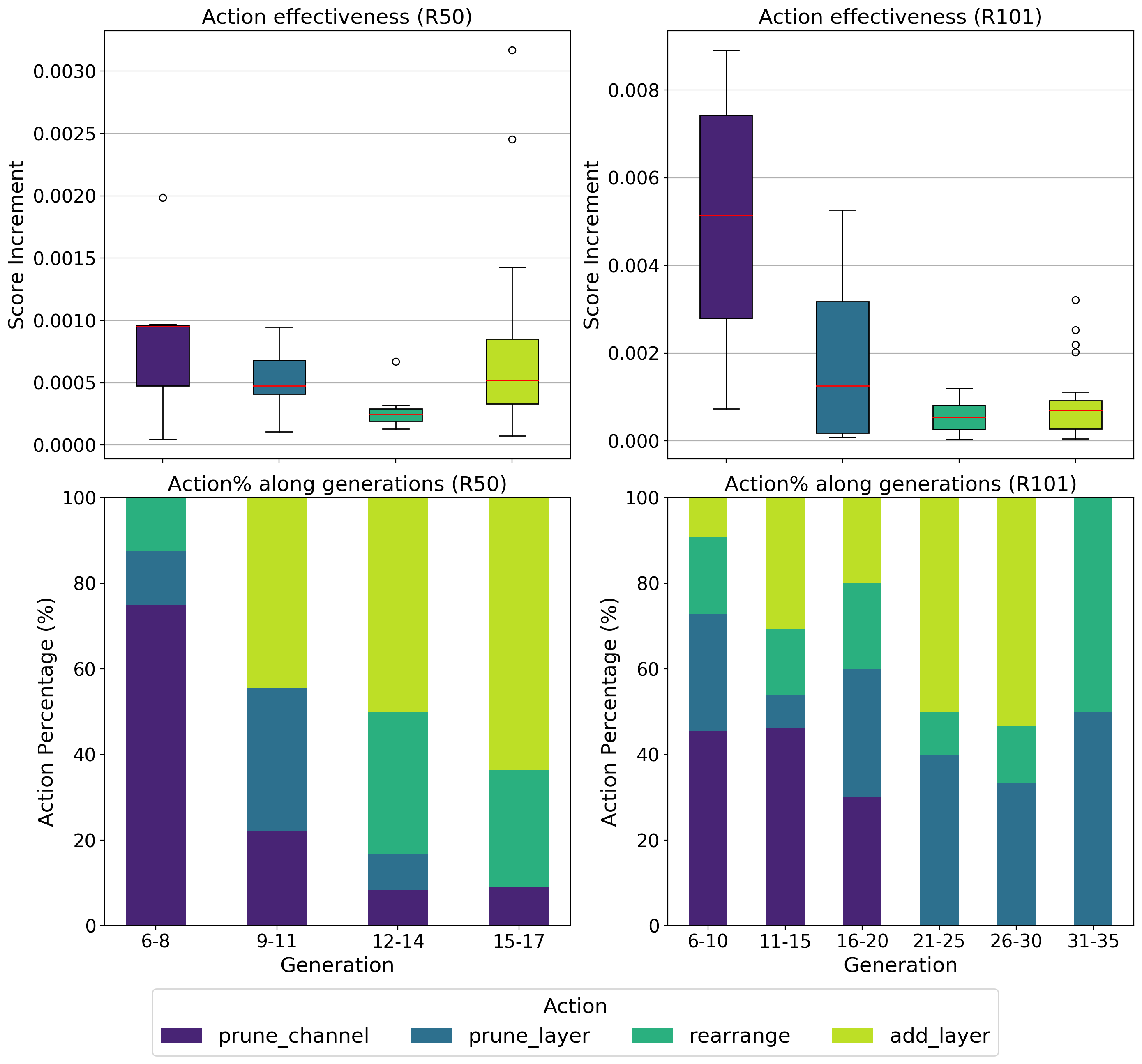}
\par\end{centering}
\vspace{-2mm}

\caption{\label{fig:action-analysis}\textbf{Top}: the overall score increment
brought by each action throughout the search; \textbf{Bottom}: percentage
of beneficial actions throughout generations. Channel pruning is mostly
adopted during early phases, and contributes most score increment.
Other actions brings more fine-grained adjustments and occur mainly
in later phases. }

\vspace{-3mm}
\end{figure}

\begin{figure}
\begin{centering}
\vspace{-2mm}
\includegraphics[scale=0.3]{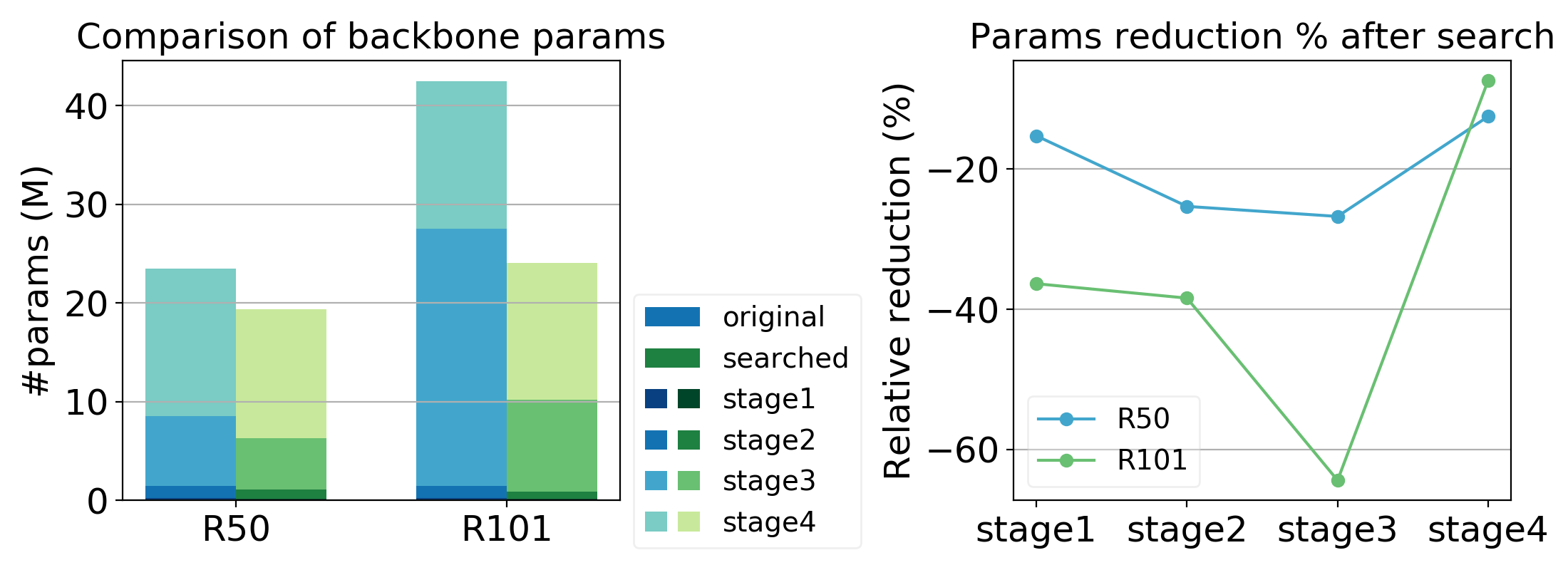}
\par\end{centering}
\vspace{-2mm}

\caption{\label{fig:backbone-analysis}The computation allocation of detector's
backbone before and after the search. For detectors with classic ResNet-based
backbones, the computation reduction is mostly allocated at stage
3, followed by stage 2 and stage 1.}

\vspace{-3mm}
\end{figure}

We study how the student evolves along the search process by analyzing
the actions improving the score function $H$ taken throughout the
generations (generation increases when student's performances is boosted)
for our R50- and R101-based search. Figure \ref{fig:action-analysis}
shows the shift of focus in balancing the performance-complexity tradeoff.
We can see that channel pruning contributes the most score increment.
In early phases, channel pruning occurs more often to adjust the network
as a whole; while in later phases, Add-layer, Prune-layer and Rearrange
follow to adjust the computation allocation at each stage in a fine-grained
manner. 

\vspace{-3mm}

\subsubsection{Computation Allocation for Detector Backbone}

In Figure \ref{fig:backbone-analysis}, we show the backbone's computation
allocation of our R50- and R101-based detectors before and after the
search. The computation at stage 3 is reduced most dramatically, followed
by stage 2 and stage 1. This implies the redundancy distribution in
manually designed ResNet models, which provides the community with
some prior knowledge for detector's backbone design.

\section{Conclusion}

This paper present a new way of jointly optimizing NAS, pruning and
KD to boost the performance and reduce the complexity of object detectors.
Extensive experiments are conducted to show the superior performance
of our proposed Joint-DetNAS framework. We believe our method has
the potential to be extended to tasks other than object detection.

\part*{Supplementary Materials}

\begin{appendix}

\section{Implementation Detail}

\thispagestyle{empty}

\subsection{NAS}

\subsubsection{Search Space}

We adopt the ResNet-based detectors as the search space due to its
popularity in the detection community. Specifically, the backbone
architecture is divided into four stages, where the feature resolution
halves and the number of output channels doubles at the beginning
of each stage. Basic block is used for R18-based students, while Bottleneck
Block is used for other students and the teacher pool. In the following
sections, ``layer'' and ``block'' are used interchangeably.

\subsubsection{Student Morphism}

The student's action spaces contains four actions: (1)\textbf{ Channel
Pruning}, (2) \textbf{Layer Pruning}, (3) \textbf{Add-Layer} and (4)
\textbf{Rearrange}.

We specify the definition of $f_{evolve}$ for each action.
\begin{itemize}
\item \textbf{Pruning} The parameters are first ranked globally by an importance
measure, then the least important ones are removed while the rest
are inherited. For\textbf{ Channel Pruning}, the importance measure
is the magnitude of each BN's channel weights. For\textbf{ Layer Pruning},
the importance measure is the parameter's L1 norm.
\item \textbf{Add-Layer} aims to introduce extra capacity into the detector
while maintain the performance of the predecessor. This is realized
by initializing the block as an identity mapping. Specifically, for
each block in ResNet whose output can represented as $H(x)=F(x)+x$,
we make $F(x)$ equal to 0 by applying Dirac initialization \cite{zagoruyko2017diracnets}
to the CONV layers and zero-initializing the last BN layer\cite{he2019bag,goyal2017accurate}.
The new layer is appended to the end of the selected stage.
\item \textbf{Rearrange}, a stage is firstly selected, then the layer at
the beginning or the end of the stage is moved to its neighboring
stage by modifying its stride, the parameters can then be directly
inherited. 
\end{itemize}

\subsubsection{Elastic Teacher Pool (ETP)}

\textbf{Subnet Space.} In our implementation of the ETP, the super-network
is set to have the same depth and 1.5x width as ResNet101. Specifically,
the depths and the width coefficients are {[}3, 4, 23, 3{]} and {[}1.5,
1.5, 1.5, 1.5{]} at each stage, respectively. During our integrated
progressive shrinking training, the subnet space is gradually expanded
to include smaller subnets. At the final phase, the smallest subnet
in the space has depths {[}2,2,2,2{]} and width coefficients {[}1.0,
1.0, 1.0, 1.0{]} at each stage, all the subnets in between can be
sampled and trained. The width coefficients can be 1.0, 1.25 or 1.5.

\textbf{Dynamic Resolution. }We use $512\times512$, $800\times600$,
$1080\times720$ and $1333\times800$ as the predefined resolutions,
from which one is randomly sampled during each training iteration.

\textbf{Phases of integrated progressive shrinking.} (1) \textbf{Training
the super-network: }the super-network is firstly trained with dynamic
resolution, which is later used as the teacher detector to distill
other subnets. (2) \textbf{First shrinking phase: }the depths and
widths of the subnet space are expanded to {[}3,4,12-23,3{]} and {[}1.25-1.5,1.25-1.5,1.25-1.5,1.25-1.5{]}
, respectively. (3)\textbf{ Second shrinking phase: }the depths and
widths of the subnet space are expanded to {[}2-3,2-4,2-23,2-3{]}
and {[}1.0-1.5,1.0-1.5,1.0-1.5,1.0-1.5{]} , respectively. During (2)
and (3), one subnet is randomly sampled from the subnet space and
trained in each training iteration. Dynamic resolution is adopted
throughout the training process.

\textbf{Training details. }The teacher pool is trained from scratch
on 32 GPUs with batch size $2\times32$ (2 for each GPU). Synchronized
BN is adopted to normalize input distribution across multiple nodes,
which addresses the issue cause by small batch size. Step learning
rate schedule is used throughout training. The initial learning rate
and training epochs for the 3 phases are described in Table \ref{tab:shrinking-setting}.

\begin{table}
\begin{centering}
\tabcolsep 0.1in{\scriptsize{}}%
\begin{tabular}{c|c|c}
\hline 
\multirow{2}{*}{Phase} & {\small{}Initial} & \multirow{2}{*}{{\small{}Epochs}}\tabularnewline
 & learning rate & \tabularnewline
\hline 
Super-net training & {\small{}0.12} & 48\tabularnewline
Shrinking phase 1 & {\small{}0.04} & 24\tabularnewline
{\small{}Shrinking phase 2} & {\small{}0.04} & {\small{}36}\tabularnewline
\hline 
\end{tabular}{\scriptsize\par}
\par\end{centering}
\caption{\label{tab:shrinking-setting}Training schedule at each phase of our
ETP.}

\vspace{-2mm}
\end{table}

\subsubsection{Details for search process\label{subsec:Details-for-search}}

The student's architecture is fixed during the first 5 search iterations
to make the search more stable. At the beginning of each search iteration,
one student-teacher pair is sampled from the topk list according to
the score ranking. The size of topk list is set to 5. In $f_{score}$,
$\beta$ is set to 0.8 for all base detector; $\alpha$ is set to
0.1 for X101 to encourage higher performance, while it is set to 0.4
for other base detectors. During fast evaluation phase, $\left\{ S_{new}^{\theta^{'}},T_{new}\right\} $
is trained for 3 epochs under cosine learning rate schedule, where
the initial learning rate is set to 0.01; the batch size is 4; synchronized
BN is adopted.

\subsection{Knowledge Distillation}

\textbf{Adaptation function. }The adaptation function $f_{adap}(\cdot)$
is implemented as a 3x3 Conv layer to match the feature dimensions
of the student-teacher pair. The output dimension is set to 256 and
the stride is set to 1.

\textbf{Proposal matching.} The student and the teacher have different
proposals, leading to unmatched outputs which cannot be directly distilled.
We solve this by sharing student's proposals with the teacher.

\subsection{Pruning}

The existence of skip connections constrain the blocks in the same
stage to have identical output dimensions. Thus, the channels can
not be arbitrarily pruned. To address this issue, the BN's weights
in projection mapping (the skip connection of the stage's first block)
are used to prune the output channel of all blocks in the stage. The
other channels inside the block are determined by the weights of the
two BN modules at the middle.

To encourage channel sparsity, we enforce a regularization term on
the weights of BN. We set the loss weight $\lambda$ to be $1\times10^{-5}$
in our implementation.

\section{Encoding of the Searched Architecture}

The student's backbone architecture is encoded as the output channels
of each convolutional layer in each block at every stage. Blocks and
stages are separated by ``-'' and ``{]}, {[}'', respectively.
We list out the encodings of students obtained with different base
detectors and the corresponding input resolutions

\textbf{R18. Student:} {[}(64, 64){]}, {[}(128, 128)-(128, 128){]},
{[}(256, 256)-(256, 256){]}, {[}(512, 512)-(512, 512); \textbf{Input
size: }$1080\times720$.

\textbf{R50. Student: }(58, 59, 205)-(60, 64, 205)-(63, 62, 205){]},
{[}(127, 128, 314)-(109, 122, 314)-(127, 123, 314)-(125, 124, 314){]},
{[}(256, 255, 591)-(243, 245, 591)-(237, 247, 591)-(243, 246, 591)-(252,
244, 591)-(252, 254, 591){]}, {[}(509, 507, 1856)-(509, 506, 1856)-(508,
507, 1856){]}; \textbf{Input size: }$1080\times720$.

\textbf{R101. Student: }{[}(49, 62, 202)-(35, 33, 202)-(56, 62, 202){]},
{[}(123, 128, 300)-(57, 90, 300)-(117, 113, 300)-(124, 117, 300){]},
{[}(255, 254, 321)-(65, 127, 321)-(32, 47, 321)-(32, 63, 321)-(120,
161, 321)-(132, 181, 321)-(162, 232, 321)-(175, 241, 321)-(143, 237,
321)-(199, 246, 321)-(210, 238, 321)-(201, 225, 321)-(210, 215, 321)-(211,
222, 321)-(201, 208, 321)-(198, 206, 321)-(220, 213, 321)-(226, 221,
321)-(234, 221, 321)-(237, 222, 321){]}, {[}(249, 229, 321)-(245,
231, 321)-(511, 478, 2031)-(507, 503, 2031)-(491, 477, 2031){]}; \textbf{Input
size: }$1080\times720$.

\textbf{X101. Student:} {[}(128, 128, 256)-(112, 112, 256)-(124, 124,
256){]}, {[}(256, 256, 512)-(256, 256, 512)-(256, 256, 512)-(256,
256, 512){]}, {[}(512, 512, 1024)-(448, 448, 1024)-(480, 480, 1024)-(496,
496, 1024)-(512, 512, 1024)-(464, 464, 1024)-(416, 416, 1024)-(416,
416, 1024)-(416, 416, 1024)-(416, 416, 1024)-(432, 432, 1024)-(496,
496, 1024)-(400, 400, 1024)-(400, 400, 1024)-(464, 464, 1024)-(464,
464, 1024)-(432, 432, 1024)-(352, 352, 1024)-(400, 400, 1024)-(384,
384, 1024)-(272, 272, 1024)-(384, 384, 1024)-(384, 384, 1024){]},
{[}(384, 384, 1024)-(352, 352, 1024)-(1024, 1024, 2048)-(864, 864,
2048)-(384, 384, 2048){]}; \textbf{Input size: }$1333\times800$.

\section{Illustration of the search process}

\begin{figure}
\begin{centering}
\vspace{-2mm}
\par\end{centering}
\begin{centering}
\includegraphics[scale=0.18]{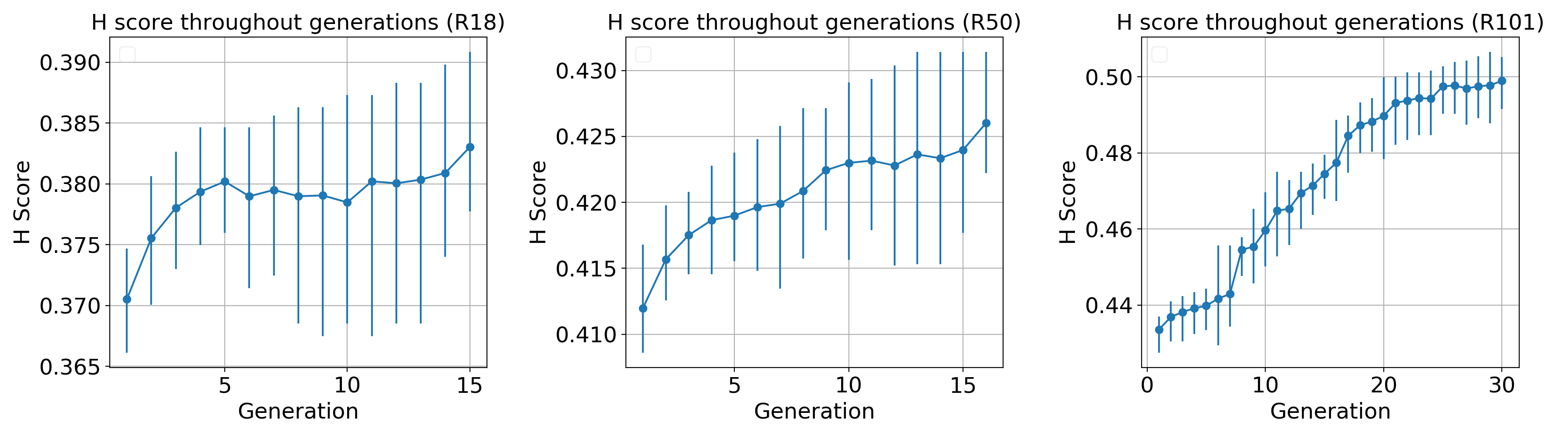}
\par\end{centering}
\begin{centering}
\vspace{-2mm}
\par\end{centering}
\caption{\label{fig:score-error}The $H$ score of sampled student detectors
throughout generation. Joint-DetNAS can consistently optimize the
performance-complexity tradeoff for various base detectors. Weight
inheritance strategy consistently improve the student's score throughout
the search.}

\centering{}\vspace{-3mm}
\end{figure}

\begin{figure}
\begin{centering}
\vspace{-2mm}
\par\end{centering}
\begin{centering}
\includegraphics[scale=0.45]{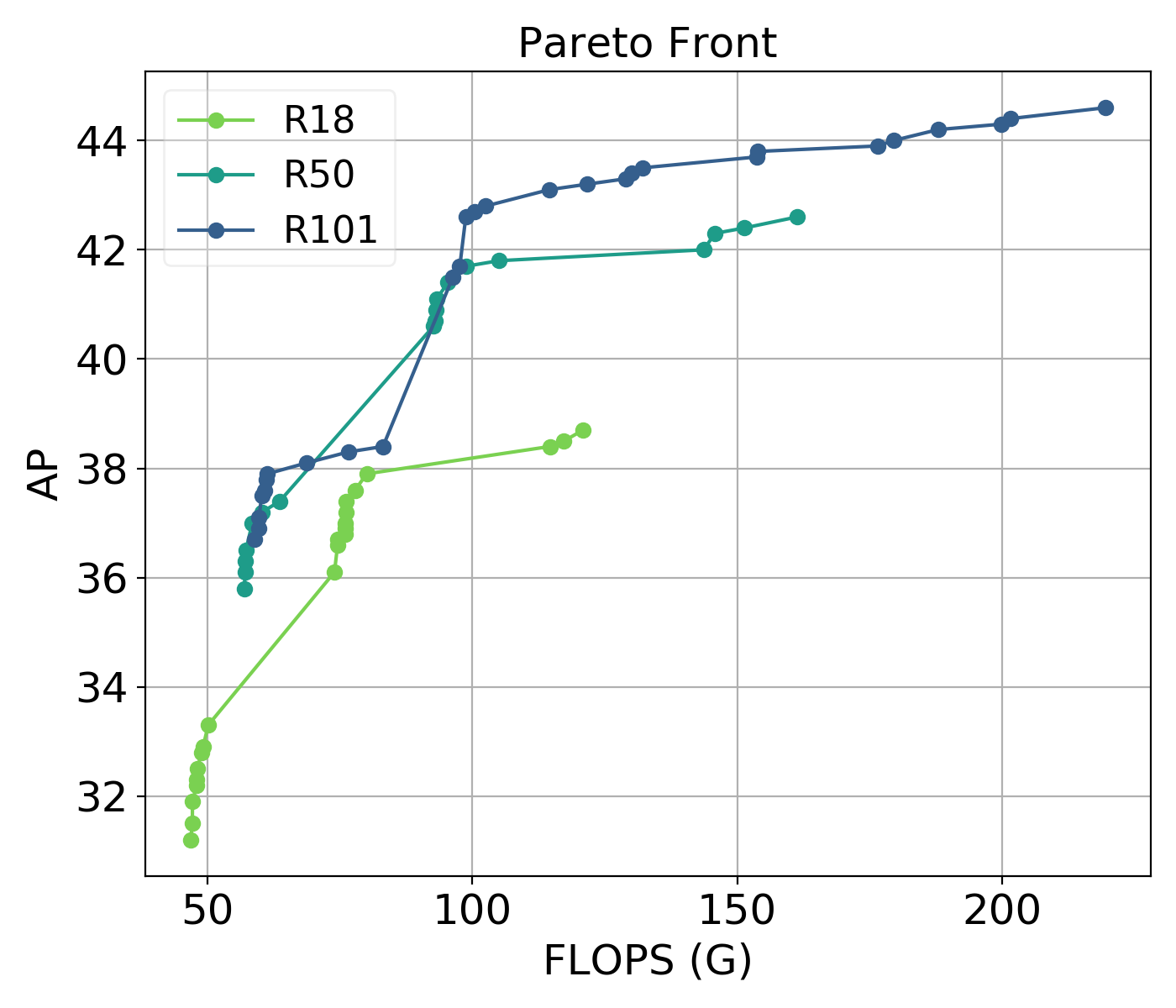}
\par\end{centering}
\begin{centering}
\vspace{-2mm}
\par\end{centering}
\caption{\label{fig:pareto}The Pareto optimal of various base detectors. As
can be seen, R101 almost dominates both R18 and R50, indicating that
given the same score function, starting with a larger base detector
can often achieve better result.}

\centering{}\vspace{-3mm}
\end{figure}

In Figure \ref{fig:score-error}, we show the $H$ score (defined
in Section 3.1.4 of the paper) of sampled student detectors throughout
generation. The results verify that Joint-DetNAS can consistently
optimize the performance-complexity tradeoff for various base detectors.
In addition, weight inheritance strategy enables the student's score
to be consistently improved throughout the search. We excluded $512\times512$
input resolution from the plot since it presents a clear performance
gap with other resolutions.

In Figure \ref{fig:pareto}, we show the Pareto optimal of various
base detectors. R101 almost dominates both R18 and R50, which indicates
that given the same score function, starting with a larger base detector
is often the better choice, because base detector with higher capacity
can be adjusted more flexibly, thus derive a better performance-complexity
tradeoff.

\section{Post-search Fine-tuning Further Improves Performance}

\begin{table}
\vspace{-2mm}

\begin{centering}
\tabcolsep 0.03in{\scriptsize{}}%
\begin{tabular}{c|c|c|c|c|c}
\hline 
{\footnotesize{}Base model} & {\footnotesize{}Group} & {\footnotesize{}Input size} & {\footnotesize{}FLOPS (G)} & {\footnotesize{}FPS} & {\footnotesize{}${\rm AP}$}\tabularnewline
\hline 
\hline 
\multirow{2}{*}{{\footnotesize{}R18-FPN}} & {\footnotesize{}baseline} & {\footnotesize{}$1333\times800$} & {\footnotesize{}160.5} & {\footnotesize{}28.2} & {\footnotesize{}36.0}\tabularnewline
 & \textbf{\footnotesize{}ours} & {\footnotesize{}$1080\times720$} & \textbf{\footnotesize{}117.3$^{-27\%}$} & \textbf{\footnotesize{}33.0$^{+17\%}$} & \textbf{\footnotesize{}38.5$\uparrow$39.8}\tabularnewline
\hline 
\multirow{2}{*}{{\footnotesize{}R50-FPN}} & {\footnotesize{}baseline} & {\footnotesize{}$1333\times800$} & {\footnotesize{}215.8} & {\footnotesize{}20.5} & {\footnotesize{}39.5}\tabularnewline
 & \textbf{\footnotesize{}ours} & {\footnotesize{}$1080\times720$} & \textbf{\footnotesize{}145.7$^{-32\%}$} & \textbf{\footnotesize{}25.4$^{+24\%}$} & \textbf{\footnotesize{}42.3$\uparrow$43.2}\tabularnewline
\hline 
\multirow{2}{*}{{\footnotesize{}R101-FPN}} & {\footnotesize{}baseline} & {\footnotesize{}$1333\times800$} & {\footnotesize{}295.7} & {\footnotesize{}15.9} & {\footnotesize{}41.4}\tabularnewline
 & \textbf{\footnotesize{}ours} & {\footnotesize{}$1080\times720$} & \textbf{\footnotesize{}153.9$^{-48\%}$} & \textbf{\footnotesize{}23.3$^{+47\%}$} & \textbf{\footnotesize{}43.9$\uparrow$44.3}\tabularnewline
\hline 
\multirow{2}{*}{{\footnotesize{}X101-FPN}} & {\footnotesize{}baseline} & {\footnotesize{}$1333\times800$} & {\footnotesize{}286.9} & {\footnotesize{}13.2} & {\footnotesize{}42.9}\tabularnewline
 & \textbf{\footnotesize{}ours} & {\footnotesize{}$1333\times800$} & \textbf{\footnotesize{}266.3$^{-7\%}$} & \textbf{\footnotesize{}14.0$^{+6\%}$} & \textbf{\footnotesize{}45.7$\uparrow$46.0}\tabularnewline
\hline 
\end{tabular}{\scriptsize\par}
\par\end{centering}
\vspace{-1mm}

\caption{\label{tab:post-finetune}The performance of found detector with post-search
fine-tuning for various input base detectors. The fine-tuning lasts
for 16 epochs; cosine learning rate schedule is adopted, with initial
learning rate set to 0.01. The value on the left and right of \textbf{$\uparrow$
}are the searched detector's performance and its fine-tuned performance,
respectively.}
\vspace{-2mm}
\end{table}

Although the obtained student detector can achieve competitive performance
without additional training, we want to show that applying post-search
fine-tuning to the student-teacher pair is able to further improve
the student's performance. The results are demonstrated in Table \ref{tab:post-finetune}.

\section{Iterative Training Does Not Hurt Performance}

\begin{figure}
\begin{centering}
\vspace{-2mm}
\par\end{centering}
\begin{centering}
\includegraphics[scale=0.18]{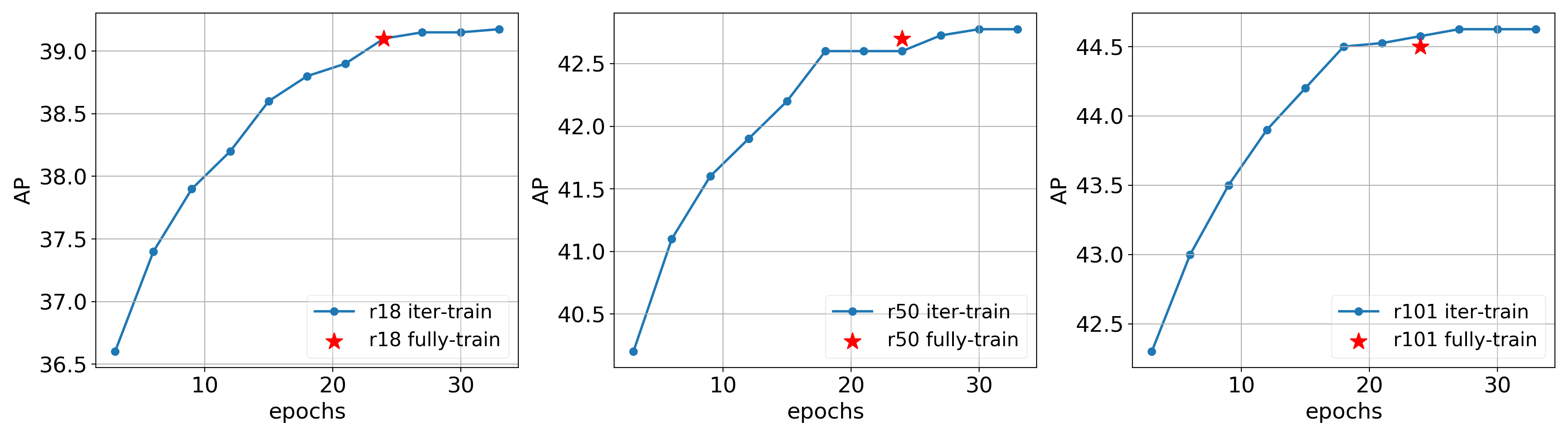}
\par\end{centering}
\begin{centering}
\vspace{-2mm}
\par\end{centering}
\caption{\label{fig:iterative_training}Comparison between iterative training
and fully training. The super-net in the ETP is used as teacher for
both iterative training and fully training. The result shows that:
(1) the convergence speeds are comparable, and (2) the final performance
of iterative training is on par with fully training.}

\centering{}\vspace{-3mm}
\end{figure}

\begin{figure*}
\begin{centering}
\vspace{-2mm}
\par\end{centering}
\begin{centering}
\includegraphics[scale=0.5]{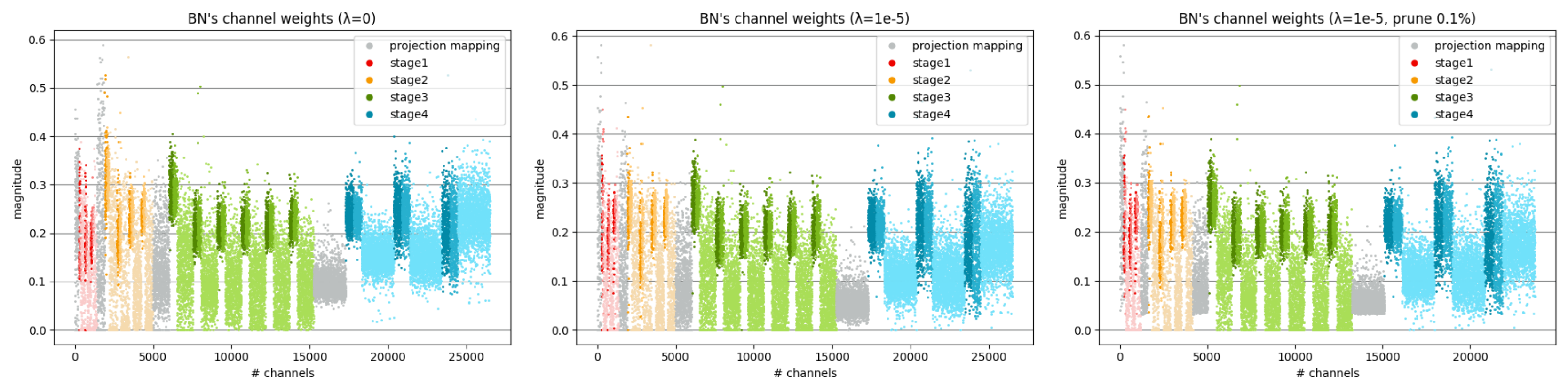}
\par\end{centering}
\begin{centering}
\vspace{-2mm}
\par\end{centering}
\caption{\label{fig:bn_prune}Analysis of BN's channel weights in the backbone.
R50-FPN is used for analysis. The three graphs demonstrate the BN's
weights of:\textbf{ Left: }normally trained detector; \textbf{Middle:
}detector trained with the regularization term, $\lambda$ = $1\times10^{-5}$;
\textbf{Right: }pruning 10\% channels from the backbone. More BN's
channel weights are close to 0 after the regularization is enforced,
which encourages sparsity.}

\centering{}\vspace{-3mm}
\end{figure*}

In the framework, the student detector is trained iteratively in each
search iteration during fast evaluation. Each iterative training process
lasts for three epochs with cosine learning rate schedule. We comparing
it with fully training in this experiment. Specifically, we fix the
student detector and use the super-net in the ETP as teacher. Then
we plot the change of AP with the training time for iterative training.
Iterative training follows the same setting as mentioned in \ref{subsec:Details-for-search}.
Fully training adopts 2x schedule and cosine learning rate decay,
the initial learning rate is 0.02. The result in Figure \ref{fig:iterative_training}
shows that: (1) the convergence speeds are comparable, and (2) the
final performance of iterative training is on par with fully training.

\section{Search with ETP }

\begin{table}
\vspace{-1mm}

\begin{centering}
\tabcolsep 0.01in{\scriptsize{}}%
\begin{tabular}{c|c|c|c|c}
\hline 
\multirow{2}{*}{{\small{}Search Method}} & \multirow{2}{*}{{\small{}FLOPS}} & \multirow{2}{*}{{\small{}${\rm AP}$}} & {\small{}\#Searched} & {\small{}Search cost}\tabularnewline
 &  &  & \multirow{1}{*}{{\small{}architectures}} & {\small{}(GPU days)}\tabularnewline
\hline 
{\small{}NAS-FPN (R50-7@256) \cite{Ghiasi_2019_CVPR}} & {\small{}281.3} & {\small{}39.9} & {\small{}10000} & {\small{}\textgreater\textgreater 500}\tabularnewline
{\small{}SP-NAS \cite{jiang2020sp}} & {\small{}349.3} & {\small{}41.7} & {\small{}200} & {\small{}200}\tabularnewline
\hline 
\textbf{\small{}ours}{\small{} (ETP-R50)} & {\small{}149.1} & {\small{}41.9} & {\small{}200} & \textbf{\small{}119}\tabularnewline
\textbf{\small{}ours}{\small{} (ETP-R101)} & {\small{}180.0} & {\small{}43} & {\small{}200} & \textbf{\small{}120}\tabularnewline
\hline 
\textbf{\small{}ours}{\small{} (Joint-DetNAS-R50)} & \textbf{\small{}145.7} & \textbf{\small{}42.3} & {\small{}100} & {\small{}185}\tabularnewline
\textbf{\small{}ours}{\small{} (Joint-DetNAS-R101)} & \textbf{\small{}153.9} & \textbf{\small{}43.9} & {\small{}100} & {\small{}200}\tabularnewline
\hline 
\end{tabular}{\scriptsize\par}
\par\end{centering}
\vspace{1mm}

\caption{\label{tab:compare-differen-search-1} Comparison between ETP search,
our Joint-DetNAS and previous works. The results demostrate that,
both ETP search and Joint-DetNAS outperform previous works: ETP search
is more efficient, while Joint-DetNAS achieves higher performance.}
\vspace{-1mm}
\end{table}

In fact, ETP can already serve as a search space, from which detectors
can be directly sampled. We compare the search result of ETP with
other NAS methods and our Joint-DetNAS in Table \ref{tab:compare-differen-search-1}.
The comparison shows that, both ETP search and Joint-DetNAS outperform
previous works: ETP search is more efficient, while Joint-DetNAS achieves
higher performance. Furthermore, the Joint-DetNAS framework is applicable
for different student architecture families without retraining the
teacher pool, thus is more flexible and economical.

\section{Ablation Study of Distillation for Object Detection}

\begin{table}
\begin{centering}
{\scriptsize{}}%
\begin{tabular}{c|c}
\hline 
{\small{}Method} & {\small{}AP}\tabularnewline
\hline 
{\small{}Baseline R18} & {\small{}34.0}\tabularnewline
\hline 
{\small{}Whole Feature \cite{chen2017learning}} & {\small{}35.2$^{+1.2}$}\tabularnewline
{\small{}Anchor Mask (fixed) \cite{wang2019distilling}} & {\small{}35.6$^{+1.6}$}\tabularnewline
{\small{}Gaussian Mask \cite{sun2020distilling}} & {\small{}35.4$^{+1.4}$}\tabularnewline
{\small{}Proposal Feature} & \textbf{\small{}36.7$^{+2.7}$}\tabularnewline
\hline 
\end{tabular}{\scriptsize\par}
\par\end{centering}
\vspace{-2mm}

\caption{\label{tab:KD-settings}Comparison between different foreground attention
mechanisms. Proposal feature outperforms the other mask based methods
by a large margin. Thus, we adopt this approach in our framework.
The student is trained under 1x schedule.}
\vspace{-2mm}
\end{table}

\begin{table}
\begin{centering}
{\footnotesize{}\tabcolsep 0.000003in}{\scriptsize{}}%
\begin{tabular}{c|c|c|c|c}
\hline 
{\small{}Proposal} & {\small{}RCNN} & \multicolumn{2}{c|}{{\small{}RCNN bbox}} & \multirow{2}{*}{{\small{}AP}}\tabularnewline
\cline{3-4} \cline{4-4} 
{\small{}Feature} & {\small{}cls} & {\small{}original} & {\small{}class-aware} & \tabularnewline
\hline 
- & - & - & - & {\small{}34.0 (R18)}\tabularnewline
- & - & - & - & {\small{}37.4 (R50)}\tabularnewline
\hline 
{\small{}$\checked$} &  &  &  & {\small{}36.7$^{+2.7}$}\tabularnewline
 & {\small{}$\checked$} &  &  & {\small{}35.8$^{+1.8}$}\tabularnewline
 &  & {\small{}$\checked$} &  & {\small{}34.8$^{+0.8}$}\tabularnewline
 &  &  & {\small{}$\checked$} & {\small{}35.7$^{+1.7}$}\tabularnewline
\hline 
 & {\small{}$\checked$} &  & {\small{}$\checked$} & {\small{}36.4$^{+2.4}$}\tabularnewline
{\small{}$\checked$} & {\small{}$\checked$} &  & {\small{}$\checked$} & \textbf{\small{}$^{\dagger}$37.9$^{+3.9}$}\tabularnewline
\hline 
\end{tabular}{\scriptsize\par}
\par\end{centering}
\vspace{-2mm}

\caption{\label{tab:KD-modules}Analysis on the effectiveness of each component
in our KD framework. The first two rows are baseline APs of R18 and
R50 FPN detectors; The student is trained under 1x schedule; \textbf{\small{}$\dagger$}
at the top left of AP indicates that the student outperforms the teacher
under the same 1x training schedule.}
\vspace{-2mm}
\end{table}

\textbf{Comparison of different ways to distill feature level information.}
Most previous detection KD methods {\small{}\cite{chen2017learning,wang2019distilling,sun2020distilling}}
aim to better distill teacher's feature level information. We compare
the mask based methods with the adopted proposal feature distillation
in Table \ref{tab:KD-settings} and found that the latter results
in the most performance gain, while being the simplest to implement.

\textbf{Analysis of each component in our KD framework. }The ablation
study of each component is shown in Table \ref{tab:KD-modules}. Our
experiments demonstrate that both feature level and prediction level
distillation bring considerable improvement. We can also see that
our proposed class-aware localization loss brings noticeable improvement
relative to the original approach which directly distill the localization
outputs. The student is R18-FPN and trained under 1x schedule, while
the teacher is trained under 2x+ms schedule.

\section{Ablation Study of Pruning for Object Detection}

We analyze the effect of the regularization term as well as the pattern
of all BNs' weights in the detector's backbone in Figure \ref{fig:bn_prune}.
As shown in the graph, more BN's channel weights are close to 0 after
the regularization is enforced. In addition, BN's weights in the third
projection mapping are smaller, thus causing the third stage to be
pruned the most. This also indicates that the third stage contains
the most redundancy.

\end{appendix}

{\small{}\bibliographystyle{ieee_fullname}
\bibliography{reference}
}{\small\par}
\end{document}